\documentclass[11pt]{article}
\usepackage{jair}
\usepackage{theapa}
\usepackage{amsmath}
\usepackage{amsfonts}
\newcommand{\LPP}{{\cal L}^{QU}}
\newcommand{\thmcolon}{\hspace{-.45em}{\bf :}}
\newtheorem{THEOREM}{Theorem}[section]
\newenvironment{theorem}{\begin{THEOREM} \thmcolon  }%
                        {\end{THEOREM}}
\newtheorem{LEMMA}[THEOREM]{Lemma}
\newenvironment{lemma}{\begin{LEMMA} \thmcolon  }%
                      {\end{LEMMA}}
\newtheorem{COROLLARY}[THEOREM]{Corollary}
\newenvironment{corollary}{\begin{COROLLARY} \thmcolon  }%
                          {\end{COROLLARY}}
\newtheorem{PROPOSITION}[THEOREM]{Proposition}
\newenvironment{proposition}{\begin{PROPOSITION} \thmcolon  }%
                            {\end{PROPOSITION}}
\newtheorem{DEFINITION}[THEOREM]{Definition}
\newenvironment{definition}{\begin{DEFINITION} \thmcolon  \rm}%
                            {\end{DEFINITION}}
\newtheorem{CLAIM}[THEOREM]{Claim}
                            {\end{CLAIM}}
\newtheorem{EXAMPLE}[THEOREM]{Example}
\newenvironment{example}{\begin{EXAMPLE} \thmcolon  \rm}%
                            {\end{EXAMPLE}}
\newtheorem{REMARK}[THEOREM]{Remark}
\newenvironment{remark}{\begin{REMARK} \thmcolon  \rm}%
                            {\end{REMARK}}

\newcommand{\thm}{\begin{theorem}}
\newcommand{\lem}{\begin{lemma}}
\newcommand{\pro}{\begin{proposition}}
\newcommand{\dfn}{\begin{definition}}
\newcommand{\rem}{\begin{remark}}
\newcommand{\xam}{\begin{example}}
\newcommand{\cor}{\begin{corollary}}
\newcommand{\prf}{\noindent{\bf Proof:} }
\newcommand{\ethm}{\end{theorem}}
\newcommand{\elem}{\end{lemma}}
\newcommand{\epro}{\end{proposition}}
\newcommand{\edfn}{\bbox\end{definition}}
\newcommand{\erem}{\bbox\end{remark}}
\newcommand{\exam}{\bbox\end{example}}
\newcommand{\ecor}{\end{corollary}}
\newcommand{\eprf}{\bbox\vspace{0.1in}}
\newcommand{\beqn}{\begin{equation}}
\newcommand{\eeqn}{\end{equation}}
\newcommand{\bbox}{\vrule height7pt width4pt depth1pt}

\newenvironment{oldthm}[1]{\medskip\par\noindent{\bf Theorem #1:} \em \noindent}{\par}
\newenvironment{oldlem}[1]{\medskip\par\noindent{\bf Lemma #1:} \em \noindent}{\par}
\newenvironment{oldcor}[1]{\medskip\par\noindent{\bf Corollary #1:} \em \noindent}{\par}
\newenvironment{oldpro}[1]{\medskip\par\noindent{\bf Proposition #1:} \em \noindent}{\par}
\newcommand{\othm}[1]{\begin{oldthm}{\ref{#1}}}
\newcommand{\eothm}{\end{oldthm} \medskip}
\newcommand{\olem}[1]{\begin{oldlem}{\ref{#1}}}
\newcommand{\eolem}{\end{oldlem} \medskip}
\newcommand{\ocor}[1]{\begin{oldcor}{\ref{#1}}}
\newcommand{\eocor}{\end{oldcor} \medskip}
\newcommand{\opro}[1]{\begin{oldpro}{\ref{#1}}}
\newcommand{\eopro}{\end{oldpro} \medskip}
\newcommand{\sat}{\models}
\newcommand{\rimp}{\Rightarrow}

\newcommand{\nthm}[1]{\begin{oldthm}{#1}}
\newcommand{\enthm}{\end{oldthm} \medskip}

\renewcommand{\phi}{\varphi}

\newcommand{\COMMENTOUT}[1]{}
\newcommand{\commentout}[1]{}
\newcommand{\compl}[1]{\ensuremath{\overline{#1}}}
\newcommand{\union}{\ensuremath{\cup}}
\newcommand{\inter}{\ensuremath{\cap}}

\newcommand{\cM}{\ensuremath{\mathcal{M}}}
\newcommand{\cP}{\ensuremath{\mathcal{P}}}
\newcommand{\cF}{\ensuremath{\mathcal{F}}}
\newcommand{\cK}{\ensuremath{\mathcal{K}}}
\newcommand{\cR}{\ensuremath{\mathbb{R}}}
\newcommand{\cN}{\ensuremath{\mathbb{N}}}

\newcommand{\upm}{\ensuremath{\upsilon}}

\newcommand{\prm}{\ensuremath{\mu}}
\newcommand{\world}{\ensuremath{\Omega}}
\newcommand{\alg}{\ensuremath{\Sigma}}

\newcommand{\truep}{\mbox{\textit{true}}}
\newcommand{\falsep}{\mbox{\textit{false}}}
\newcommand{\true}{\mbox{\textbf{true}}}
\newcommand{\false}{\mbox{\textbf{false}}}
\newcommand{\intension}[1]{[\![ #1 ]\!]}
\newcommand{\eventM}[1]{\ensuremath{\intension{#1}_M}}
\newcommand{\eventMp}[1]{\ensuremath{\intension{#1}_{M'}}}

\newcommand{\ax}{\textbf{AX$^{up}$}}
\newcommand{\axiom}[1]{\textbf{#1}}
\newcommand{\siff}{\ensuremath{\Leftrightarrow}}
\newcommand{\sep}{\ensuremath{~:~}}
\newcommand{\om}{\{\!\{}
\newcommand{\cm}{\}\!\}}
\jairheading{17}{2002}{57-81}{12/01}{8/02}
\ShortHeadings{A Logic for Reasoning about Upper
Probabilities}{Halpern \& Pucella}
\firstpageno{57}

\title{A Logic for Reasoning about Upper Probabilities}
\author{\name Joseph Y. Halpern \email halpern@cs.cornell.edu\\
\name Riccardo Pucella \email riccardo@cs.cornell.edu\\
\addr Department of Computer Science\\
Cornell University\\
Ithaca, NY 14853\\
http://www.cs.cornell.edu/home/halpern}

\begin{document}
\maketitle

\begin{abstract} 
We present a propositional logic 
to reason about the uncertainty of events, where the uncertainty is
modeled by a set 
of probability measures assigning an interval of probability to each
event. We give a sound and complete axiomatization for the logic, and
show  that the satisfiability problem is NP-complete, no harder than
satisfiability for propositional logic.   
\end{abstract}

\section{Introduction}

Various measures exist that attempt to quantify uncertainty. For 
many trained in the use of
probability theory, probability measures are an obvious
choice. However, probability cannot easiliy capture
certain situations of interest. 
Consider a simple example: suppose we have a 
bag of 100 marbles; we know 30 are red and we know the remaining 70
are either blue or yellow,  although we do not know the exact
proportion of blue and yellow. If we are modeling the situation where
we pick a ball from the bag at random, we need to assign a probability 
to three different events: picking up a red ball (\textit{red-event}),
picking up a blue ball (\textit{blue-event}), and picking up a yellow
ball (\textit{yellow-event}).  We can clearly assign a
probability of .3 to \textit{red-event},
but there is no
clear probability to assign to \textit{blue-event} or
\textit{yellow-event}. 

One way to approach this problem is to
represent the uncertainty using a set of probability measures, with a probability
measure for each possible proportion of blue and yellow
balls. For instance, we could use the set of probabilities
$\cP=\{\prm_\alpha\sep\alpha\in[0,.7]\}$, where $\prm_\alpha$ gives
\textit{red-event} 
probability $.3$, \textit{blue-event} probability $\alpha$, and
\textit{yellow-event} probability $.7-\alpha$. 
To any set of probabilities \cP{} we can assign a pair 
of functions, the upper and lower probability measure,
that for an event $X$ give the supremum (respectively, the
infimum) of the probability of $X$ according to the probability
measures in \cP. These measures can be used to deal with uncertainty in 
the manner described above, where the lower and upper probability of
an event defines a range of probability for that 
event.%
\footnote{Note that using sets of probability measures is not the only
way to model this situation. An alternative approach, using inner
measures, is studied by Fagin and Halpern \citeyear{Fagin91}.}
(This example can be viewed as giving a frequentist interpretation of
upper probabilities.  Upper probabilities can also be given a subjective
interpretation, for example, by considering the odds at which someone
would be willing to accept or reject a bet \cite{Smith,Walley91}.)

Given a measure of uncertainty, one can define a 
logic for reasoning about it.
Fagin, Halpern and
Megiddo \citeyear{Fagin90} 
(FHM from now on) 
introduce a logic for reasoning about 
probabilities, with a possible-worlds semantics that assigns a
probability to each possible world. They provide an axiomatization
for the logic, which they prove sound and complete with respect to the
semantics. 
They also show that the satisfiability problem for the
logic, somewhat surprisingly, is NP-complete, 
and hence no harder than the satisfiability problem for propositional logic.
They moreover show how their logic can be extended to other notions of
uncertainty, such as inner measures \cite{Fagin91} and Dempster-Shafer
belief functions \cite{Shafer76}. 

In this paper, we describe a logic for reasoning about upper
probability measures, along the lines of the FHM logic.
The logic allows reasoning about linear inequalities involving
upper probabilities measures. 
Like the logics considered in FHM, our logic 
is agnostic as to
the interpretation of upper probabilities, whether frequentist or
subjectivist. 
The main 
challenge
is to derive a provably complete axiomatization of the 
logic;
to do this,
we need a characterization of upper probability measures in terms of
properties 
that can be expressed in 
the logic.
Many 
semantic characterizations of upper probability measures have
been proposed in the 
literature. The characterization 
of 
Anger and Lembcke \citeyear{Anger85} turns out to be best suited for
our purposes. 
Even though we are reasoning about potentially infinite sets of
probability measures, the satisfiability problem for our logic remains
NP-complete. 
Intuitively, we need
guess only
a small number of probability measures to satisfy any given
formula, polynomially many in the size of the formula. Moreover,
these probability measures can be taken to be defined on a finite state
space, again polynomial in the size of the formula. 
Thus, we need to basically
determine polynomially many values---a value for each probability
measure at each state---to decide the satisfiability of a
formula. 

The rest of this paper is structured as follows. In Section \ref{s:prelim}, we
review the required material from probability theory and the theory of 
upper probabilities. In Section \ref{s:logic}, we present the logic
and an axiomatization. In Section~\ref{s:compl}, we prove that the
axiomatization is sound
and complete with respect to the natural semantic models expressed in
terms of upper probability spaces. 
Finally, in Section~\ref{s:decision}, we prove that the
decision problem for the logic is NP-complete. 
The proofs of the 
new,
more technical results are given in Appendix~\ref{a:proofs}.
To make the paper self-contained, we also review Anger and Lembcke's
results in Appendix~\ref{a:upm}.
\section{Characterizing Upper Probability Measures}
\label{s:prelim}

We start with a brief review of the relevant definitions.
Recall 
that a
probability measure is a 
function $\prm:\alg\rightarrow[0,1]$ for
$\alg$ an algebra of subsets of $\world$ (that is $\alg$ is
closed under complements and unions), satisfying
$\prm(\emptyset)=0$, $\prm(\world)=1$, and $\prm(A\cup
B)=\prm(A)+\prm(B)$ for 
all disjoint sets $A,B$ in $\Sigma$.%
\footnote{If $\world$ is infinite, we could also require that $\alg$ be
a $\sigma$-algebra (i.e., closed under countable unions) and that 
$\mu$ 
be  countably additive.  
Requiring
countable additivity would not affect our results, since we show that
we can take $\world$ to be finite. 
For ease of exposition, we have not required it.} 
A probability space is a tuple
$(\world,\alg,\prm)$, where $\world$ is a set, $\alg$ is an
algebra of subsets of $\world$ (the measurable sets), and
$\prm$ is a probability measure defined on 
$\alg$. 
Given a set $\cP$ of probability measures, let
$\cP^*$ be the 
upper probability measure
 defined by $\cP^*(X) = \sup \{\prm(X)\sep
\prm\in\cP\}$ for $X\in\alg$.%
\footnote{
  In the literature, the term upper probability is 
sometimes
  used in a more restricted sense than here. For example, Dempster
  \citeyear{Dempster67} uses the term to denote a class of measures which
  were later characterized as Dempster-Shafer belief functions
  \cite{Shafer76};  
  belief functions are in fact upper probability measures
  in our sense, but the converse 
  is not true \cite{Kyb}. 
In the measure theory
  literature, what we call upper probability measures 
are a special case of
  \emph{upper envelopes} of measures, 
which 
are defined as the sup
  of sets of general measures, not just probability measures. 
}
Similarly, 
$\cP_*(X)=\inf\{\prm(X)\sep \prm\in\cP\}$ is the lower probability
of $X\in\alg$. A straightforward derivation shows that the
relationship $\cP_*(X)=1-\cP^*(\compl{X})$ holds between upper and
lower probabilities,
where $\compl{X}$ is the complement of $X$ in $\world$. 
 Because of this duality, we restrict the 
discussion to upper probability measures in this paper, with the
understanding that results for lower probabilities can be 
similarly derived. Finally, 
an \emph{upper probability space} is a tuple
$(\world,\alg,\cP)$ where $\cP$ is a set of probability measures on
$\alg$.  

We would like a set of properties that completely
characterizes upper probability measures. 
In other words, we would like a set of properties that allow us to
determine if a function $f:\alg\rightarrow\cR$ (for an algebra
$\alg$ of subsets of $\world$) is an upper probability measure, that
is, whether there exists a set $\cP{}$ of probability measures such
that for all $X\in\alg$, $\cP^*(X)=f(X)$.%
\footnote{It is possible to define a notion of upper probability over
an arbitrary set of subsets of $\world$, not necessearily an algebra, by
simply requiring that $f$ 
coincides with $\cP^*$ on its domain, for some set $\cP$ of
probability measures. See Walley \citeyear{Walley91} for details.}

One approach to the characterization of upper probability measures
is to adapt the characterization of Dempster-Shafer belief functions;
these functions are 
known to be the lower envelope of the probability measures that dominate them,
and thus form a subclass of the class of lower probability
measures. 
By the duality noted earlier, 
a characterization of lower probability measures would yield a
characterization of upper probability measures.
The characterization of belief
functions is derived from a generalization of the 
following
inclusion-exclusion principle for probabilities 
(obtained by replacing the equality with an inequality):
\[ \mu(\bigcup_{i=1}^{n}A_i) = \sum_{i=1}^{n}
(-1)^{i-1}(\sum_{\substack{J\subseteq\{1,\ldots,n\}\\ |J|=i}}\mu(\bigcap_{j\in
J}A_j)). \]  

It seems reasonable that a characterization of lower (or upper)
probability measures could be derived along similar lines. 
However,
as is well known, most properties derivable from the
inclusion-exclusion principle  
(which include most of
the properties 
reported in the literature)
are insufficient to characterize upper probability measures.
Huber \citeyear[p. 257]{Huber81} and Walley \citeyear[p. 85]{Walley91} 
give examples showing the insufficiencies of such properties. 
\commentout{
Our investigation 
into characterizing upper probability measures started from axioms
directly derived through an application of the inclusion/exclusion
principle. (These properties
refer to both upper and lower probabilities for convenience, but they
can easily be recast using only one of the measures, 
by exploiting the duality between
upper and lower probability measures). 
Such properties have been reported, for example, by Walley
\cite{Walley91}. 
}

To give a sense of the insufficiency of simple properties, consider
the following ``inclusion-exclusion''--style properties, some of which 
are
taken from \cite{Walley91}. 
To simplify  
the statement of
these properties, let $\cP^{-1} = \cP^*$ and $\cP^{+1} = \cP_*$.
\begin{itemize}
\item[(1)] $\cP^*(A_1\union\cdots\union A_n) \leq
\sum_{i=1}^{n}\sum_{|I|=i}(-1)^{i+1}\cP^{(-1)^i}(\bigcap_{j\in
I}A_j)$,
\item[(2)] $\cP_*(A_1\union\cdots\union A_n) \geq
\sum_{i=1}^{n}\sum_{|I|=i}(-1)^{i+1}\cP^{(-1)^{i+1}}(\bigcap_{j\in 
I}A_j)$,
\item[(3)] $\cP_* (A\union B) + \cP_* (A\inter B) \leq \cP_* (A) + \cP^* (B) \leq \cP^*(A\union B) + \cP^* (A\inter B)$,
\item[(4)] $\cP_* (A) + \cP_* (B) \leq \cP_* (A\union B) + \cP^* (A\inter B) \leq \cP^*(A) + \cP^* (B)$,
\item[(5)] $\cP_* (A) + \cP_* (B) \leq \cP_* (A\inter B) + \cP^* (A\union B) \leq \cP^*(A) + \cP^* (B)$.
\end{itemize}

Note that without the alternation between upper probabilities and lower
probabilities, (1) and (2) would just be the standard notions of
subadditivity and superadditivity, respectively.  While subadditivity
and superadditivity hold 
for
upper and lower probabilities, respectively, (1) and
(2) are stronger properties.
It is easily verified that all five properties hold for upper
probability measures. The question is whether they completely
characterize the class of upper probability measures. We show the
inherent incompleteness of these properties by 
proving
that they are all derivable from the following simple property, 
which is by
itself insufficient to characterize upper probability measures: 
\begin{itemize} 
\item[(6)] 
If
$A\inter B=\emptyset$, then $\cP^*(A) 
+ \cP_*(B) \leq \cP^*(A\union B) \leq \cP^*(A)+\cP^*(B)$. 
\end{itemize} 

\pro
\label{p:prop6}
Property (6) implies properties (1)-(5).
\epro
Observe that our property (6) is already given by Walley
\citeyear[p. 84]{Walley91}, as properties (d) and (e).
The following example shows the insufficiency of Property (6).
Let $\cP$
be the set of probability measures $\{\mu_1,\mu_2,\mu_3,\mu_4\}$ over
$\world=\{a,b,c,d\}$ (with $\alg$ containing all subsets of $\world$)
defined on singletons by
\[\begin{array}{cccc}
  \mu_1 (a) = \frac{1}{4} & \mu_1(b)=\frac{1}{4} & \mu_1(c)=\frac{1}{4} & \mu_1(d) = \frac{1}{4} \\[1em]
  \mu_2(a) = 0 & \mu_2(b) = \frac{1}{8} & \mu_2(c)=\frac{3}{8} & \mu_2(d) = \frac{1}{2} \\[1em]
  \mu_3(a) = \frac{1}{8} & \mu_3(b)=\frac{3}{8} & \mu_3(c)=0 & \mu_3(d) = \frac{1}{2} \\[1em]
  \mu_4(a) = \frac{3}{8} & \mu_4(b)=0 & \mu_4(c) = \frac{1}{8} &
  \mu_4(d) = \frac{1}{2}, 
  \end{array}
\]
and extended by additivity to all of $\alg$. This defines an upper
probability measure $\cP^*$ over $\alg$. Consider the function
$\upm_\epsilon:\alg\rightarrow[0,1]$ defined by 
\[ \upm_\epsilon (X) = \left\{ \begin{array}{ll}
   \cP^*(X)+\epsilon & \mbox{if $X=\{a,b,c\}$}\\
   \cP^*(X) & \mbox{otherwise.}
                               \end{array}\right. \]
We claim that the function $\upm_\epsilon$, for small enough
$\epsilon>0$, satisfies property (6), but cannot be an upper
probability measure.
\pro
\label{p:notupm}
For $0<\epsilon<\frac{1}{8}$, the function $\upm_\epsilon$ satisfies
property (6), but is not an upper probability measure. That
is, we cannot find a set $\cP'$ of probability measures such that
$\upm_\epsilon=(\cP')^*$. 
\epro

This example 
clearly illustrates 
the need to go beyond the inclusion-exclusion principle to find
properties 
that characterize 
upper probability measures. As it turns out,
various complete characterizations have been described in the
literature 
\cite{Lorentz52,Huber76,Huber81,Williams76,Wolf77,Giles82,Anger85,Walley91}.
Most of these characterizations are obtained by considering
upper and lower \emph{expectations}, rather than working directly with
upper and lower probabilities.
Anger and Lembcke \citeyear{Anger85} give a characterization in terms of
upper and lower probabilities.  Since their characterization is
particularly well-suited to the logic presented in the next section, we
review it here.

The characterization is based on the notion of
{\em set cover}.  A set $A$ is said to be \emph{covered
$n$ times} by a multiset $\om A_1,\ldots,A_m \cm$ of sets if every
element of $A$ appears in at least $n$ 
sets from $A_1,\ldots,A_m$:~for
all $x\in A$, there exists 
distinct
$i_1,\ldots,i_n$ in $\{1,\ldots,m\}$ such
that for all $j\leq n$, $x\in A_{i_j}$. 
It is important to note here that $\om A_1, \ldots, A_m \cm$ is a multiset,
not a set; the $A_i$'s are not
necessarily distinct.  (We use the $\om \, \cm$ notation to denote
multisets.) 
An {\em $(n,k)$-cover of $(A,\world)$\/} is a multiset $\om A_1, \ldots,
A_m \cm$ that covers $\world$ $k$ times and covers $A$ $n+k$ times.
For example, $\{\{1,2\}, \{2,3\}, \{1,3\}\}$ covers $\{1,2,3\}$ 2 times,
and $\{\{\{1,2\}, \{2,3\}, \{1,3\}, \{2\}, \{2\}\}\}$ is a (2,2) cover
of $(\{2\},\{1,2,3\})$.

The notion of $(n,k)$-cover is the key 
concept
in Anger and Lembcke's characterization of upper probability measures.

\thm \label{t:upm}
{\rm \cite{Anger85}}
Suppose that $\world$ is a set, $\alg$ is an algebra of subsets of
$\world$, and $\upm:\alg\rightarrow\cR$. Then there
exists a set $\cP$ of probability measures with $\upm=\cP^*$ if and
only if $\upm$ satisfies the following three properties: 
\begin{quote}
\begin{itemize}

\item[\textbf{{\rm UP1}}.] $\upm (\emptyset) = 0$,
\item[\textbf{{\rm UP2}}.] $\upm (\world) = 1$,
\item[\textbf{{\rm UP3}}.] for all natural numbers $m,n,k$ and all subsets
$A_1, \ldots, A_m$ in $\alg$, if $\om A_1, \ldots, A_m \cm$ is an
$(n,k)$-cover of $(A,\world)$, then $k+n\upm(A)\leq\sum_{i=1}^{m}\upm(A_i)$.
\end{itemize}
\end{quote}
\ethm
\prf
We reproduce a proof of this result in Appendix~\ref{a:upm}. 
\eprf

Note that \textbf{UP1} is redundant in the presence of \textbf{UP2}
and \textbf{UP3}. Indeed, $\om \world,\emptyset \cm$ is a
$(0,1)$-cover of $(\world,\world)$, and applying \textbf{UP3} yields
$\upm(\emptyset)+\upm(\world)=1$. Since \textbf{UP2} states that 
$\upm(\world)=1$, this means that $\upm(\emptyset)=0$. A further
consequence of \textbf{UP3} is that if $A\subseteq B$, then
$\upm(A)\le\upm(B)$, since $\om B\cm$ is a $(1,0)$-cover of
$(A,\world)$.  Therefore, for all $A\in\alg$, $\upm(A)\in[0,1]$. 

We need to strengthen Theorem~\ref{t:upm} in order to prove the main
result of this paper, namely, the completeness of the axiomatization
of the logic we introduce in the next section. We show that if the
cardinality of the state space 
$\world$
is finite, then we need only finitely many instances of property
\textbf{UP3}. Notice that we cannot derive this from Theorem
\ref{t:upm} alone:~even if $|\world|$ is finite, \textbf{UP3} does not
provide any bound on $m$, the number of sets to consider in an $(n,k)$
cover of a set $A$.  Indeed, there does not
seem to be any \textit{a priori} reason why the value of $m$, $n$, and $k$ can
be bounded. Bounding this value of $m$ (and hence of $n$ and $k$, since
they are no larger than $m$) is 
one of the key technical results of this paper, 
and a necessary foundation for our work.

\thm
\label{t:fupm}
There exist constants $B_0,B_1,\ldots$ such that 
if $\world$ is a finite set, $\alg$ is an algebra of subsets of
$\world$, and $\upm:\alg\rightarrow\cR$, 
then there exists a set $\cP$ of
probability measures such that $\upm=\cP^*$ if and only if $\upm$
satisfies the following properties:
\begin{quote}
\begin{itemize}
\item[\textbf{{\rm UPF1}}.] $\upm (\emptyset) = 0$,
\item[\textbf{{\rm UPF2}}.] $\upm (\world) = 1$,
\item[\textbf{{\rm UPF3}}.] 
for all integers $m, n, k \le B_{|\world|}$ and all sets $A_1, \ldots,
A_m$, if
$\om A_1, \ldots, A_m \cm$ is an $(n,k)$-cover of $(A,\world)$,
then $k+n\upm(A)\leq\sum_{i=1}^{m}\upm(A_i)$.
\end{itemize}
\end{quote}
\ethm

Property \textbf{UPF3} is significantly weaker than \textbf{UP3}. In
principle, checking that \textbf{UP3} holds for a given function
requires checking that it holds for 
arbitrarily large collections of sets,
even if the underlying set $\world$ is finite. On the other
hand, \textbf{UPF3} guarantees that 
if $\world$ is finite, then
it is in fact sufficient to look
at collections of size at most $B_{|\world|}$. This observation is key
to 
the completeness result.

Theorem~\ref{t:fupm} does not prescribe any values for the constants
$B_0,B_1,\ldots$. Indeed, the proof found in Appendix~\ref{a:proofs}
relies on a Ramsey-theoretic argument that does not even provide a
bound on the $B_i$'s. We could certainly attempt to obtain such
bounds, but obtaining them is completely unnecessary for our
purposes.  To get completeness 
of the axiomatization of the logic introduced in the next section,
it is sufficient for there to exist finite constants $B_0,B_1,\ldots$.

\section{The Logic}
\label{s:logic}

The syntax for the logic is straightforward, and is taken
from FHM. We fix a set $\Phi_0=\{p_1,p_2,\ldots\}$
of \emph{primitive propositions}. The set $\Phi$ of
\emph{propositional formulas} is the closure of $\Phi_0$ under
$\wedge$ and $\neg$. We assume a special propositional formula
\truep, and abbreviate $\neg\truep$ as
\falsep. We use $p$ to represent primitive propositions, and
$\phi$ and $\psi$ to represent propositional formulas. A \emph{term}
is an expression of the form $\theta_1 l(\phi_1) + \cdots+\theta_k
l(\phi_k)$, where $\theta_1,\ldots,\theta_k$ are reals and
$k\geq 1$. A \emph{basic likelihood formula} is a statement of the form
$t\geq\alpha$, where $t$ is a term and $\alpha$ is a real. A
\emph{likelihood formula} is a Boolean combination of basic likelihood
formulas. 
We use $f$ and $g$ to represent likelihood formulas. 
We use obvious
abbreviations where needed, such as $l(\phi)-l(\psi)\geq a$ for
$l(\phi)+(-1)l(\psi)\geq a$, $l(\phi)\geq l(\psi)$ for
$l(\phi)-l(\psi)\geq 0$, $l(\phi)\leq a$ for $-l(\phi)\geq -a$,
$l(\phi)< a$ for $\neg(l(\phi)\geq a)$ and $l(\phi)=a$ for
$(l(\phi)\geq a)\wedge(l(\phi)\leq a)$. 
Define the length $|f|$
of the likelihood formula $f$ to be the number of symbols required to
write $f$, where each coefficient is counted as one symbol.
Let $\LPP$ be the language consisting of likelihood formulas.
(The QU stands for {\em quantitative uncertainty}.  The name for the
logic is taken from \cite{Hal34}.)

In FHM, the operator $l$ was interpreted as either ``probability'' or
``belief'' (in the sense of Dempster-Shafer).  Under the first
interpretation, a formula such as $l(\phi) + l(\psi) \ge 2/3$ would be
intereted as ``the probability of $\phi$ plus the probability of $\psi$
is at least $2/3$''.  Here we interpret $l$ as upper probaiblity.  Thus,
the logic allows us to make statements about inequalities involving
upper probabilities.

To capture this interpretation,
we assign a semantics to formulas in $\LPP$ using an upper
probability space, as defined in Section \ref{s:prelim}. Formally, an
\emph{upper probability structure} is 
a tuple $M=(\world,\alg,\cP,\pi)$ where $(\world,\alg,\cP)$ is an
upper probability space and $\pi$ associates with each state (or world) in
$\world$ a truth assignment on the primitive propositions in
$\Phi_0$. Thus, $\pi(s)(p)\in\{\true,\false\}$ for
$s\in\world$ and $p\in\Phi_0$. Let $\eventM{p} = \{s\in\world \sep
\pi(s)(p)=\true\}$. We call $M$ \emph{measurable} if for each
$p\in\Phi_0$, $\eventM{p}$ is measurable. If $M$ is measurable then
$\eventM{\phi}$ is measurable for all propositional formulas
$\phi$. In this paper, we restrict 
our attention to measurable upper probability structures. Extend
$\pi(s)$ to a truth assignment on all propositional formulas in a
standard way, and associate with each propositional formula the set
$\eventM{\phi}=\{s\in\world \sep \pi(s)(\phi)=\true\}$. An easy
structural induction shows that $\eventM{\phi}$ is a measurable set. If
$M=(\world,\alg,\cP,\pi)$, let
\[\begin{array}{l}
   M \models \theta_1 l(\phi_1) + \cdots + \theta_k l(\phi_k)\geq\alpha \mbox{~~iff~~} 
     \theta_1\cP^*(\eventM{\phi_1}) + \cdots + \theta_k\cP^*(\eventM{\phi_k})\geq\alpha \\
   M \models \neg f \mbox{~~iff~~} M \not\models f\\
   M \models f\wedge g \mbox{~~iff~~} M \models f \mbox{~and~} M \models g.
  \end{array}\]

Note that $\LPP$ can express lower probabilities:~it follows from
the duality between upper and lower probabilities that $M\models
-l(\neg\phi)\geq\beta-1$ iff $\cP_*(\eventM{\neg\phi})\geq\beta$.\footnote{Another approach, more in keeping
with FHM, would be to interpret $l$ as a lower
probability measure. On the other hand,
interpreting $l$ as an upper probability measure is more in keeping with the
literature on upper probabilities.}

Consider the following axiomatization \ax\ 
of upper probability,
which we prove sound and complete in the next section. 
The key axioms  are simply a
translation into $\LPP$ of the 
characterization of upper probability given in Theorem~\ref{t:upm}.
As in FHM, \ax\ is divided into three parts, 
dealing respectively
with propositional reasoning, reasoning about linear inequalities, and
reasoning about upper probabilities.
\begin{itemize}
\item[]\textit{Propositional reasoning}
\item[]\axiom{Taut}. All instances of propositional tautologies
in $\LPP$ (see below).
\item[]\axiom{MP}. From $f$ and $f\implies g$ infer $g$.
\item[]\textit{Reasoning about linear inequalities}
\item[]\axiom{Ineq}. All instances of valid formulas about linear
inequalities (see below).
\item[]\textit{Reasoning about upper probabilities}
\item[]\axiom{L1}. $l(\falsep)=0$.
\item[]\axiom{L2}. $l(\truep)=1$.
\item[]\axiom{L3}. $l(\phi)\geq 0$.
\item[]\axiom{L4}. $l(\phi_1)+\cdots+l(\phi_m)-nl(\phi)\geq k$ if 
$\phi\rimp\bigvee_{J \subseteq \{1, \ldots, m\},\,
|J|=k+n} \bigwedge_{j\in J}\phi_j$ and\\
$\bigvee_{J \subseteq \{1, \ldots, m\},\,
|J|=k} \bigwedge_{j\in J} \phi_j$ are propositional tautologies.%
\footnote{Note that, according to the syntax of $\LPP$, $\phi_1, \ldots,
\phi_m$ must be propositional formulas.}
\item[]\axiom{L5}. $l(\phi)=l(\psi)$ if $\phi\siff\psi$ is a
propositional tautology.
\end{itemize}
The only difference between \ax\ and the axiomatization for reasoning
about probability given in FHM is that the axiom 
$l(\phi \land \psi) + l(\phi \land \neg \psi) = l(\phi)$ in
FHM, which expresses the additivity of probability, is
replaced by \axiom{L4}.  Although it may not be immediately obvious,
\axiom{L4} is the logical analogue of 
\textbf{UP3}.  
To see this, first note that 
$\om A_1,\ldots,A_m \cm$ covers $A$ $m$ times if and only if
$A\subseteq\bigcup_{J\subseteq\{1,\ldots,m\},\, |J|=n}\bigcap_{j\in
J}A_j$.  Thus, 
the formula $\phi\rimp\bigvee_{J\subseteq\{1,\ldots,m\},\, |J|=k+n}\bigwedge_{j\in J}\phi_j$
says that $\phi$ (more precisely, the set of worlds where $\phi$ is
true) is covered $k+n$ times by $\om \phi_1, \ldots, \phi_n \cm$, while 
$\bigvee_{J\subseteq\{1,\ldots,m\},\, |J|=k}\bigwedge_{j\in J}\phi_j$ says that the whole
space is covered $k$ times by $\om \phi_1, \ldots, \phi_n \cm$; 
roughly speaking, 
the multiset
$\om \phi_1, \ldots, \phi_n \cm$ is an $(n,k)$-cover of $(\phi,\truep)$.  The
conclusion of \axiom{L4} thus corresponds to the conclusion of
\textbf{UP3}. 
Note that in the same way that \textbf{UP1} follows from \textbf{UP2}
and \textbf{UP3}, axiom \axiom{L1} (as well as \axiom{L3}) follows
from \axiom{L2} and \axiom{L4}. 

Instances of \axiom{Taut} include all formulas of the form $f\vee\neg f$, 
where $f$ is an arbitrary formula in $\LPP$.
We could replace \axiom{Taut} by a
simple collection of axioms that characterize propositional reasoning
(see, for example, \cite{Men}), but we have chosen to focus on aspects
of reasoning about upper probability. 

As in FHM,
the axiom \textbf{Ineq} includes ``all valid formulas about linear
inequalities.'' An \emph{inequality formula} is a formula
of the form $a_1 x_1 + \dots + a_n x_n \geq c$, over variables
$x_1,\ldots,x_n$. 
An inequality formula is said to \emph{valid} if it is true under every
possible assignment of real numbers to variables. To get an instance
of \axiom{Ineq}, we replace each variable $x_i$ that occurs in
a valid inequality formula by a primitive likelihood
term of the form $l(\phi_i)$ 
(naturally
each occurence of the
variable $x_i$ must be replaced by the same primitive likelihood term
$l(\phi_i)$). 
As with \axiom{Taut}, we can replace \axiom{Ineq}
by a sound and complete axiomatization for Boolean combinations of
linear inequalities. One such axiomatization is given in
FHM.

\COMMENTOUT{
\item[\axiom{2}.] $(\theta_1 l(\phi_1)+\cdots+\theta_k
l(\phi_k)\geq\alpha)\siff(\theta_1 l(\phi_1)+\cdots+\theta_k l(\phi_k)
+ 0l(\phi_{k+1})\geq\alpha)$ (adding and deleting 0 terms)
\item[\axiom{3}.] $(\theta_1 l(\phi_1)+\cdots+\theta_k
l(\phi_k)\geq\alpha)\rimp(\theta_{j_1} l(\phi_{j_1}) +
\cdots+\theta_{j_k} l(\phi_{j_k}\geq\alpha)$, if $j_1,\ldots,j_k$ is a 
permutation of $1,\ldots,k$ (permutation)
\item[\axiom{4}.] $(\theta_1 l(\phi_1)+\cdots+\theta_k
l(\phi_k)\geq\alpha)\wedge(\theta'_1 l(\phi_1) + \cdots + \theta'_k
l(\phi_k)\geq\alpha') \rimp
(\theta_1+\theta'_1)l(\phi_1)+\cdots+(\theta_k+\theta'_k)l(\phi_k)\geq(\alpha+\alpha')$ 
(addition of coefficients)
\item[\axiom{5}.] $(\theta_1 l(\phi_1)+\cdots+\theta_k
l(\phi_k)\geq\alpha) \rimp (\gamma\theta_1
l(\phi_1)+\cdots+\omega\theta_k l(\phi_k)\geq\gamma\alpha)$ if
$\gamma\geq 0$ (multiplication of coefficients)
\item[\axiom{6}.] $(t\geq\alpha)\vee(t\leq\alpha)$ if $t$ is a term (dichotomy)
\item[\axiom{7}.] $(t\geq\alpha)\rimp(t>\beta)$ if $t$ is a term
and $\alpha>\beta$ (monotonicity)
}

\section{Soundness and Completeness}
\label{s:compl}

A formula $f$ is \emph{provable in an axiom system \textbf{AX}} 
if $f$ can be proven using the axioms and rules of inferences of \textbf{AX}.
\textbf{AX} is \emph{sound} with respect to a class $\cM$ of structures
if every formula provable in \textbf{AX} is valid in $\cM$
(i.e., valid in every structure in $\cM$); 
\textbf{AX} is \emph{complete} with respect to $\cM$ 
if every formula valid in $\cM$ is provable in \textbf{AX}.

Our goal is to prove that \ax\ is a sound and complete axiomatization
for reasoning about upper probability 
(i.e., with respect to upper probability structures).
The soundness of \ax\ is immediate from our earlier disscussion.
Completeness is, as usual, harder.  
Unfortunately, the standard technique for proving completeness in modal
logic, which involves considering maximal consistent sets and canonical
structures (see, for example, \cite{Popkorn}) does not work.  We briefly
review the approach, just to point out the difficulties.

The standard approach uses the following definitions.
A formula $\sigma$ is 
{\em consistent\/} with an axiom system {\bf AX} if $\neg
\sigma$ is not provable from {\bf AX}.
A finite set of formulas $\{\sigma_1,\ldots,\sigma_n\}$ is
consistent 
with {\bf AX}
if the formula $\sigma_1\wedge\cdots\wedge \sigma_n$ is consistent with {\bf
AX}; an infinite
set of formulas is consistent with {\bf AX} if all its finite subsets are
consistent with {\bf AX}. 
$F$ is a {\em maximal\/} {\bf AX}-consistent set if $F$ is consistent
with {\bf AX} and no strict superset of $F$ is consistent with {\bf AX}.
If {\bf AX} includes \axiom{Taut} and \axiom{MP}, then it is not hard to
show, using only propositional reasoning,
that every {\bf AX}-consistent set of formulas
can be extended to a maximal {\bf AX}-consistent set. 
To show 
that {\bf AX} is complete with respect to some class $\cM$ of structures,
we must show that every formula that is valid in 
${\cal M}$ is provable in {\bf AX}.
To do this, it is sufficient
to show that every {\bf AX}-consistent
formula is satisfiable in some structure in ${\cal M}$. 
Typically, this is done by constructing what is called a {\em canonical
structure\/} $M^c$ in ${\cal M}$ whose 
states are the maximal {\bf AX}-consistent sets, and then showing that a
formula $\sigma$ is satisfied 
in a world $w$ in $M^c$
iff $\sigma$ is one of the formulas in the canonical set associated
with world $w$. 

Unfortunately, this approach 
cannot be used to prove completeness here. To see this, consider the
set of formulas
\[F' = \{l(\phi)\leq \frac{1}{n} ~,~ n=1,2,\ldots\} \cup \{l(\phi)> 0\}.\]
This set is clearly \ax--consistent according to our definition, since
every finite subset is satisfiable in an upper probability structure and
\ax\ is sound with respect to upper probability structures.  It thus can be 
extended to a maximal \ax--consistent set $F$.
However, the set $F'$ of formulas is not satisfiable:  
it is not possible to assign 
$l(\phi)$ a value that will satisfy all the formulas at the same
time. Hence, $F$ is not satisfiable. 
Thus, the canonical model approach, at least applied naively, simply
will not work. 

We take a different approach here, 
similar to the one taken in FHM.
We do not try to construct a single canonical model.  
Of course, we still must show that if a formula $f$ is \ax-consistent
then it is satisfiable in an upper probability structure. 
We do this by an explicit construction, depending on $f$.
We proceed as follows.

By a simple argument, we
can easily reduce the problem to  the case where $f$ is a conjunction
of basic likelihood formulas
and negations of basic likelihood formulas.
Let $p_1,\ldots,p_N$ be the primitive
propositions that appear in $f$. 
Observe that there are
$2^{2^N}$ inequivalent propositional formulas 
over
$p_1,\ldots,p_N$.
The argument goes as follows. Let
an {\em atom over $p_1, \ldots, p_N$\/} be a
formula of the form $q_1 \land \ldots \land q_N$, where $q_i$ is either
$p_i$ or $\neg p_i$.  There are clearly $2^N$ atoms over $p_1, \ldots,
p_N$.  Moreover, it is easy to see that any formula over $p_1, \ldots,
p_N$ can be written in a unique way as a disjunction of atoms.  
There are $2^{2^N}$ such disjunctions, so the claim follows.

Continuing with the construction of a structure satisfying $f$,
let $\rho_1,\ldots,\rho_{2^{2^N}}$ be some canonical listing of the
inequivalent formulas over $p_1, \ldots, p_N$.
Without loss of
generality, we assume that $\rho_1$ is equivalent to $\truep$, and
$\rho_{2^{2^N}}$ is equivalent to $\falsep$. 
Since every propositional formula over $p_1,\ldots,p_N$ is
provably equivalent to some $\rho$, it follows that $f$ is
provably equivalent to a formula $f'$ where each conjunct of $f'$ is
of the form $\theta_1 l(\rho_1)+\cdots+\theta_{2^{2^N}}
l(\rho_{2^{2^N}})
\ge \beta$.  Note that the negation of such a formula has the form
$\theta_1 l(\rho_1)+\cdots+\theta_{2^{2^N}}
l(\rho_{2^{2^N}}) < \beta$ or, equivalently, 
$(-\theta_1) l(\rho_1)+\cdots+(-\theta_{2^{2^N}})
l(\rho_{2^{2^N}}) > -\beta$.  Thus,
the formula $f$ gives rise in a natural way to a system of
inequalities of the form:
\begin{equation}
\label{eqcomp}
\begin{array}{ccc}
\theta_{1,1}l(\rho_1) + \cdots + \theta_{1,2^{2^N}}l(\rho_{2^{2^N}})
& \geq & \alpha_1\\ 
\vdots & \vdots & \vdots\\ 
\theta_{r,1}l(\rho_1)+ \cdots + \theta_{r,2^{2^N}}l(\rho_{2^{2^N}}) &
\geq & \alpha_r\\
\theta'_{1,1}l(\rho_1) + \cdots + \theta'_{1,2^{2^N}}l(\rho_{2^{2^N}})
& > & \beta_1\\
\vdots & \vdots & \vdots\\
\theta'_{s,1}l(\rho_1)+\cdots + \theta'_{s,2^{2^N}}l(\rho_{2^{2^N}}) & 
> & \beta_s.
\end{array}
\end{equation}

We can express (\ref{eqcomp}) as a 
conjunction of inequality formulas,  by replacing each occurrence
of $l(\rho_i)$ in (\ref{eqcomp}) by $x_i$.  Call this inequality formula
$\overline{f}$. 

If $f$ is satisfiable in some upper probability structure $M$, then we can
take $x_i$ to be the upper probability of $\rho_i$ in $M$; 
this gives a solution of
$\overline{f}$. 
However, 
$\overline{f}$ may have a solution without $f$ being
satisfiable.  For example, if $f$ is the formula $l(p) = 1/2 \land
l(\neg p) = 0$, then $\overline{f}$ has an obvious 
solution; $f$, however, is not 
satisfiable in an upper probability 
structure, because 
the upper probability of the
set corresponding to $p$
and the upper probability of the set corresponding to $\neg p$ must
sum to at 
least 1 in all upper probability structures.
Thus, we must add further constraints to the solution to force it to act
like an upper probability. 

\textbf{UP1}--\textbf{UP3} or, equivalently, the axioms
\textbf{L1}--\textbf{L4}, describe exactly what additional constraints
are needed.  The constraint corresponding to \textbf{L1} (or
\textbf{UP1}) is just $x_1 = 0$, since we have assumed $\rho_1$ is the
formula $\falsep$.  Similarly, the constraint corresponding to
\textbf{L2} is $x_{2^{2^N}} = 1$.  The constraint corresponding to
\textbf{L3} is $x_i \ge 0$, for $i = 1, \ldots, 2^{2^N}$.  What about
\textbf{L4}?  This seems to require an infinite collection of
constraints, just as \textbf{UP3} does.%
\footnote{Although we are dealing with only finitely many formulas here,
$\rho_1, \ldots, \rho_{2^{2^N}}$, recall that the formulas $\phi_1,
\ldots, \phi_m$ in \axiom{L4} need not be distinct, so there are
potentially infinitely many instances of \axiom{L4} to deal with.}

This
is where \textbf{UPF3}
comes into play.  It turns out that, if $f$ is satisfiable at all, it is
satisfiable in a structure with at most $2^N$ worlds, one for each atom
over $p_1, \ldots, p_N$.  Thus, we need to add only instances of
\textbf{L4} where $k, m, n < B_{2^N}$ and $\phi_1, \ldots, \phi_m, \phi$
are all among $\rho_1, \ldots, \rho_{2^{2^N}}$.  Although this is a
large number of formulas (in fact, we do not know exactly how large,
since it depends on $B_{2^N}$, which we have not computed), it suffices
for our purposes that it is a finite number. 
For each of these instances of \axiom{L4}, there is an
inequality of the form $a_1 x_1 + \cdots + a_{2^{2^N}}x_{2^{2^N}} \ge k$.
Let $\hat{f}$, the {\em inequality formula corresponding to $f$}, be the
conjunction consisting of $\overline{f}$,
together with all the inequalities 
corresponding to the relevant instances of \axiom{L4}, and the equations
and inequalities $x_1 = 0$, $x_{2^{2^N}}=1$, and $x_i \ge 0$ for $i =
1, \ldots, 2^{2^N}$, corresponding to axioms \axiom{L1}--\axiom{L3}.
\pro
\label{l:compl}
The formula $f$ is satisfiable in an upper probability structure iff the
inequality formula $\hat{f}$ has a solution.
Moreover, if $\hat{f}$ has a solution, then $f$ is satisfiable in an
upper probability structure with at most $2^{|f|}$ worlds.
\epro
\thm
\label{t:sandc}
The axiom system \ax\ is sound and complete for upper probability structures.
\ethm

\prf For soundness, it is easy to see
that every axiom is valid for upper probability structures, including
\axiom{L4}, which represents \textbf{UP3}. 

For completeness, we proceed as 
in the discussion 
above. Assume that
formula $f$ is not satisfiable in an upper probability structure; we
must show that $f$ is 
\ax--inconsistent. We first reduce $f$ to a canonical form. Let
$g_1\vee\cdots\vee g_r$ be a disjunctive normal form expression for
$f$ (where each $g_i$ is a conjunction of basic likelihood formulas
and their negations). Using propositional reasoning, we can show that
$f$ is provably equivalent to this disjunction. Since $f$ is
unsatisfiable, 
each $g_i$ must also be unsatisfiable.
Thus, it is sufficient to show that any 
unsatisfiable
conjunction of basic likelihood formulas and their negations 
is inconsistent.
Assume that $f$ is such a conjunction.
Using propositional reasoning and axiom \axiom{L5}, 
$f$ is equivalent to a likelihood formula $f'$ that refers to 
$2^{2^N}$ propositional formulas, say
$\rho_1,\ldots,\rho_{2^{2^N}}$. Since $f$ is unsatisfiable, so is
$f'$. By Proposition~\ref{l:compl},
the inequality formula $\hat{f'}$ corresponding to $f'$ has no solution.
Thus, by \axiom{Ineq}, the formula $\neg f''$ that results by replacing each
instance of $x_i$ in $\hat{f'}$ by $l(\rho_i)$ is \ax--provable.  All the
conjuncts of $f''$ that are instances of axioms \axiom{L1}--\axiom{L4}
are \ax--provable.
It follows that $\neg f'$ is \ax--provable, and hence so
is $\neg f$.  
\eprf

\section{Decision Procedure}
\label{s:decision}

Having settled the issue of the soundness and completeness of the
axiom system \ax, we turn to the problem of the complexity of deciding 
satisfiability. Recall the problem of satisfiability: given a
likelihood formula $f$, we want to determine if 
there exists an upper probability structure $M$ such that $M\models
f$. As we now show, the satisfiability problem is NP-complete, and
thus no harder than satisfiability for propositional logic.

For the decision problem to make sense, we need to 
restrict our language slightly.
If we allow real numbers as coefficients in likelihood
formulas, we have to carefully discuss the issue of representation of
such numbers. To avoid these complications, we restrict our language
(in this section)
to allow only integer coefficients. Note that we can still express
rational coefficients by the standard trick of ``clearing the
denominator''. For example, we can  
express  $\frac{2}{3}l(\phi)\geq 1$ by $2l(\phi)\geq 3$ and 
$l(\phi)\geq\frac{2}{3}$ by $3l(\phi)\geq 2$. 
Recall that we defined $|f|$ to be the length of $f$, that is, the number of symbols 
required to write $f$, where each coefficient is counted as one
symbol. 
Define
$||f||$ to be the length of the longest coefficient appearing in $f$,
when written in binary. The size of a rational number $\frac{a}{b}$,
denoted $||\frac{a}{b}||$,
where $a$ and $b$ are relatively prime, is defined to be $||a||+||b||$. 

A preliminary result required for the analysis of the decision
procedure shows that  if a formula is satisfied in some upper
probability structure, then it is satisfied in a structure 
$(\world,\alg,\cP,\pi)$, 
which is ``small''
in terms of the number of states in $\world$, the cardinality of
the set $\cP$ of probability measures, and 
the  size of the coefficients in $f$.  

\thm
\label{t:smallmodel}
Suppose $f$ is a likelihood formula that is satisfied in some upper
probability structure. Then $f$ is satisfied in a structure
$(\world,\alg,\cP,\pi)$, 
where $|\world| \le |f|^2$, $\alg = 2^{\world}$ 
(every
subset of $\world$ is measurable), $|\cP|\leq|f|$, 
$\mu(w)$ is a rational
number such that $||\mu(w)||$ is
$O(|f|^2||f||+|f|^2\log(|f|))$ for every world $w \in \world$ and
$\mu \in \cP$,
and $\pi(w)(p) = \false$ for every world $w \in \world$ and every primitive
proposition $p$ not appearing in $f$. 
\ethm

\thm
\label{t:npcompl}
The problem of deciding whether a likelihood formula is satisfiable in 
an upper probability structure is NP-complete.
\ethm
\prf
For the lower bound, it is clear that a given propositional formula
$\phi$ is satisfiable iff the likelihood formula $l(\phi)>0$ is
satisfiable, 
therefore
the satisfiability problem is NP-hard.
For the upper bound, given a likelihood formula $f$, we
guess a ``small'' satisfying structure $M=(\world,\alg,\cP,\pi)$ for $f$ 
of the form guaranteed to exist by Theorem~\ref{t:smallmodel}. 
We can describe such a model $M$ in size polynomial in $|f|$ and
$||f||$.  (The fact that $\pi(w)(p) = \false$ for every world $w \in
\world$ and every primitive proposition $p$ not appearing in $f$ means
that we must describe $\pi$ only for propositions that appear in $f$.)
We verify
that $M\models f$ as follows. Let $l(\psi)$ be an arbitrary likelihood
term in $f$. 
We compute
$\eventM{\psi}$ by checking the truth assignment of each $s\in\world$
and seeing whether this truth assignment makes $\psi$ true. 
We then replace each occurence of $l(\psi)$ in
$f$ by $\max_{\mu\in\cP}\{\sum_{s\in \eventM{\psi}}\mu(s)\}$ and verify that
the resulting expression is true. 
\eprf

\section{Conclusion} 
We have considered a logic with the same syntax as the logic for
reasoning about probability, inner measures, and belief presented in
FHM,
with uncertainty interpreted as the upper probability
of a set of probability measures.  Under this interpretation, we have
provided a sound and complete axiomatization for the logic.
We further showed that the satisfiability problem is NP-complete 
(as
it is for reasoning about probability, inner measures, and beliefs),
despite having to deal with probability structures with possibility
infinitely many states and infinite sets of probability measures.
The key step in
the axiomatization involves finding a 
characterization of upper probability measures that can be captured in
the logic.  The key step in the complexity result involves showing
that 
if a formula is satisfiable at all, it is satisfiable in a ``small''
structure, where 
the size of the state space, as well as the size of the set of
probability measures  and the size of all probabilities involved, 
are
polynomial in the length of the formula. 

Given the similarity in spirit of the results for the various
interpretations of the uncertainty operator (as a probability, inner
measure, belief function, and upper probability), 
including the fact that the complexity of the decision problem 
is NP-complete in all cases, 
we conjecture that there is some underlying result from which all
these results should follow. It would be interesting to make that
precise.

In FHM, conditional probabilities as well as probabilities are investigated.
We have not, in this paper,
discussed conditional upper probabilities. The main reason for this is
that, unlike probability, we cannot characterize conditional upper
probabilities in terms of (unconditional) upper probabilities. Thus,
our results really tell us nothing about conditional upper
probabilities. 
It might be of interest to consider a logic that allows conditional
upper probabilities as primitive likelihood terms (that is, allows
likelihood terms of the form $l(\phi \, | \, \psi)$).  While there is
no intrinsic difficult giving semantics to such a language, it is far
from clear what an appropriate axiomatization would be, or the effect
of this extension on complexity. 

Finally, it is worth noting that the semantic framework developed here
and in FHM is in fact rich 
enough to talk about gambles (that is, real-valued functions over the
set of states) and the expectation of such gambles. Expectation
functions can be defined for the different measures of uncertainty,
including upper probabilities, and it is not difficult to extend the
FHM logic in order to reason about expectation. One advantage of
working with expectation functions is that they are typically easier
to characterize than the corresponding measures; for instance, the
characterization of expected upper probabilities is much simpler than
that of upper probabilities 
\cite{Huber81,Walley,Walley91}.
However, getting a complete
axiomatization is quite nontrivial. We refer the reader to
\cite{HalPuc02:UAI} for more details on this subject.   
We remark that Wilson and
Moral \citeyear{Wilson94} take as 
their starting point Walley's notion of lower and upper previsions.
They consider when acceptance of one set of gambles implies
acceptance of another gamble.  Since acceptance involves expectation, it
cannot be expressed in the logic considered in this paper; however, it
can be expressed easiliy in the logic of \cite{HalPuc02:UAI}.

\acks{
A preliminary version of this paper appears in \emph{Uncertainty in
Artificial Intelligence, Proceedings of the Seventeenth Conference},
2001. 
Thanks to Dexter Kozen, Jon Kleinberg, and
Hubie Chen for discussions concerning set covers. 
Vicky Weissman read a draft of this paper and provided numerous
helpful comments. 
We also thank the anonymous UAI and JAIR reviewers for their useful comments
and suggestions. 
This work was supported in part by NSF under
grants IRI-96-25901 and IIS-0090145, and ONR under 
grants N00014-00-1-03-41, N00014-01-10-511, and N00014-01-1-0795.
The first author was also supported in part by a Guggenheim and a
Fulbright Fellowship while on sabbatical leave; sabbatical support from
CWI and the Hebrew University of Jerusalem is also gratefully acknowledged.
}

\appendix

\newcommand{\tg}{\ensuremath{\tilde{g}}}

\section{Proofs}\label{a:proofs}

\opro{p:prop6}
Property (6) implies properties (1)-(5).
\eopro
\prf
We introduce the following auxiliary properties to help derive the
implications:
\begin{itemize}
\item[(7)] $\cP_*(A)+\cP_*(B)\leq\cP_*(A\union B)+\cP^*(A\inter B)$.
\item[(8)] $\cP_*(A)+\cP_*(B)\leq\cP_*(A\inter B)+\cP^*(A\union B)$.
\item[(9)] $\cP_*(A\union B)+\cP_*(A\inter B)\leq\cP_*(A)+\cP^*(B)$.
\item[(10)] 
If $A\inter B=\emptyset$, then
$$\cP_*(A)+\cP_*(B)\leq\cP_*(A\union
B)\leq\cP_*(A)+\cP^*(B)\leq\cP^*(A\union B)\leq\cP^*(A)+\cP^*(B).$$
\end{itemize}
Using these properties, we show the following chain of implications:
\[\begin{array}{ccc}
  (6) \implies (10) & ~~~~~
  \begin{array}{c}
    (10) \implies (9) \implies (3)\\
    (10) \implies (7) \implies (4)\\
    (10) \implies (8) \implies (5)
  \end{array} ~~~~~ & 
  (4),(5) \implies (1),(2).
  \end{array}
\]

The implication (4), (5) $\implies$ (1), (2) follows easily by mutual
induction on 
$n$. The base case is the following instances of properties (4) and
(5):  $\cP_*(A\union B)\geq \cP_*(A)+\cP_*(B)-\cP^*(A\inter B)$ and
$\cP^*(A\union B)\leq \cP^*(A)+\cP^*(B)-\cP_*(A\inter B)$.
The details are left to the reader. 

We now prove the remaining implications.
\begin{itemize}
\item[](9) $\implies$ (3): Since (9) is already one of the inequalities in
(3), it remains to show that it implies the other inequality in (3),
that is, $\cP_*(A)+\cP^*(B)\leq\cP^*(A\union B)+\cP^*(A\inter B)$.
\begin{eqnarray*}
\cP^*(A\union B)+\cP^*(A\inter B) & = & 1-\cP_*(\compl{A\union B})+1-\cP_*(\compl{A\inter B})\\
 & = & 1-\cP_*(\compl{A}\inter\compl{B})+1-\cP_*(\compl{A}\union\compl{B})\\
 & = & 2-(\cP_*(\compl{A}\inter\compl{B})+\cP_*(\compl{A}\union\compl{B}))\\
 & = & 2-(\cP_*(\compl{B}\inter\compl{A})+\cP_*(\compl{B}\union\compl{A}))\\
 & \geq & 2-(\cP_*(\compl{B})+\cP^*(\compl{A}))\\
 & = & 1-\cP_*(\compl{B})+1-\cP^*(\compl{A})\\
 & = & \cP^*(B)+\cP_*(A).
\end{eqnarray*}

\item[](7) $\implies$ (4): Since (7) is already one of the inequalities in
(4), it remains to show that it implies the other inequality in (4),
that is, $\cP_*(A\union B)+\cP^*(A\inter B)\leq\cP^*(A)+\cP^*(B)$.
\begin{eqnarray*}
\cP^*(A)+\cP^*(B) & = & 1-\cP_*(\compl{A})+1-\cP_*(\compl{B})\\
 & = & 2-(\cP_*(\compl{A})+\cP_*(\compl{B}))\\
 & \geq & 2-(\cP_*(\compl{A}\union\compl{B})+\cP^*(\compl{A}\inter\compl{B}))\\
 & = & 1-\cP_*(\compl{A}\union\compl{B})+1-\cP^*(\compl{A}\inter\compl{B})\\
 & = & 1-\cP_*(\compl{A\inter B})+1-\cP^*(\compl{A\union B})\\
 & = & \cP^*(A\inter B)+\cP_*(A\union B).
\end{eqnarray*}

\item[](8) $\implies$ (5): Since (8) is already one of the inequalities in
(5), it remains to show that it implies the other inequality in (5),
that is, $\cP_*(A\inter B)+\cP^*(A\union B)\leq\cP^*(A)+\cP^*(B)$.
\begin{eqnarray*}
\cP^*(A)+\cP^*(B) & = & 1-\cP_*(\compl{A})+1-\cP_*(\compl{B})\\
 & = & 2-(\cP_*(\compl{A})+\cP_*(\compl{B}))\\
 & \geq & 2-(\cP_*(\compl{A}\inter\compl{B})+\cP^*(\compl{A}\union\compl{B}))\\
 & = & 1-\cP_*(\compl{A}\inter\compl{B})+1-\cP^*(\compl{A}\union\compl{B})\\
 & = & 1-\cP_*(\compl{A\union B})+1-\cP^*(\compl{A\inter B})\\
 & = & \cP^*(A\union B)+\cP_*(A\inter B).
\end{eqnarray*}
\end{itemize}

For the next implications, given $A,B$, let $Z=A\inter B$.
\begin{itemize}
\item[](10) $\implies$ (9):
\begin{eqnarray*}
\cP_*(A\union B) & = & \cP_*((A-Z)\union B)\\
& \leq & \cP_*(A-Z)+\cP^*(B) \mbox{~~~~~[since $(A-Z)\inter B = \emptyset$]}\\
 & \leq & \cP_*((A-Z)\union Z)-\cP_*(Z)+\cP^*(B)\\
 & = & \cP_*(A) + \cP^*(B) - \cP_*(A\inter B).
\end{eqnarray*}

\item[](10) $\implies$ (7):
\begin{eqnarray*}
\cP_*(A\union B) & = & \cP_*((A-Z)\union B)\\
 & \geq & \cP_*(A-Z)+\cP_*(B)\\
 & \geq & \cP_*((A-Z)\union Z) - \cP^*(Z) + \cP_*(B)\\
 & = & \cP_*(A) + \cP_*(B) - \cP^*(A\inter B).
\end{eqnarray*}

\item[](10) $\implies$ (8):
\begin{eqnarray*}
\cP^*(A\union B) & = & \cP^*((A-Z)\union B)\\
 & \geq & \cP^*(A-Z)+ \cP_*(B)\\
 & \geq & \cP_* ((A-Z)\union Z) - \cP_*(Z) + \cP_*(B)\\
 & = & \cP_*(A) + \cP_*(B) - \cP_*(A\inter B).
\end{eqnarray*}

\item[](6) $\implies$ (10): Again, since (6) already comprises two of the inequalities in
(10), it remains to show that it implies the other two, 
that is, if $A\inter B=\emptyset$, then
$$\cP_*(A)+\cP_*(B) \leq \cP_*(A\union B)\leq
\cP^*(A)+\cP_*(B).$$
First, we show that $\cP_*(A)+\cP_*(B)\leq\cP_*(A\union B)$. Using
(6), we know that
$$\cP^*(\compl{A}\inter\compl{B})+\cP_*(A)\leq
\cP^*((\compl{A}\inter\compl{B})\union A) = \cP^*(\compl{B}).$$ 
In other words,
$\cP^*(\compl{A}\inter\compl{B})\leq\cP^*(\compl{B})+\cP_*(A)$. From
this, we derive that
\begin{eqnarray*}
\cP_*(A\union B) & = & 1 - \cP^*(\compl{A\union B}) \\
 & = & 1 - \cP^*(\compl{A}\inter\compl{B}) \\
 & \geq & 1 - (\cP^*(\compl{B}) - \cP_*(A)) \\
 & = & 1 - \cP^*(\compl{B}) + \cP_*(A) \\
 & = & \cP_*(B) + \cP_*(A).
\end{eqnarray*}
Second, we show that $\cP_*(A\union B)\leq\cP^*(A)+\cP_*(B)$. Using 
(6), we know that 
$$\cP^*(\compl{A}\inter\compl{B})+\cP^*(A)\geq
\cP^*((\compl{A}\inter\compl{B})\union A) = \cP^*(\compl{B}).$$ 
(The last equality follows from 
the fact that
$(\compl{A}\inter\compl{B})\union A =
\compl{B}$ when $A \inter B = \emptyset$.)
In other words,
$\cP^*(\compl{A}\inter\compl{B})\geq\cP^*(\compl{B})-\cP^*(A)$. From
this, we derive that
\begin{eqnarray*}
\cP_*(A\union B) & = & 1 - \cP^*(\compl{A\union B}) \\ 
 & = & 1 - \cP^*(\compl{A}\inter\compl{B}) \\
 & \leq & 1 - (\cP^*(\compl{B}) - \cP^*(A)) \\
 & = & 1 - \cP^*(\compl{B}) + \cP^*(A) \\
 & = & \cP_*(B) + \cP^*(A).
~~~\eprf
\end{eqnarray*}
\end{itemize}

\opro{p:notupm}
For $0<\epsilon<\frac{1}{8}$, the function $\upm_\epsilon$ satisfies
property (6), but is not an upper probability measure. That
is, we cannot find a set $\cP'$ of probability measures such that
$\upm_\epsilon=(\cP')^*$. 
\eopro
\prf
We are given $0<\epsilon<\frac{1}{8}$. It is easy to check
mechanically that $\upm_\epsilon$ satisfies (6).  

We now show that there is no
set $\cP'$ such that
$\upm_\epsilon=(\cP')^*$. By way of contradiction, assume there is such a 
$\cP'$. By the properties of sup, this means that there is a
$\mu\in\cP'$ such that $\mu(\{a,b,c\})>\frac{3}{4}$, since
$\upm_\epsilon(\{a,b,c\})=\frac{3}{4}+\epsilon>\frac{3}{4}$. Consider
this $\mu$ in detail. Since $\mu\in\cP$, we must have for all
$X\in\alg$, $X\not=\{a,b,c\}$, that
$\mu(X)\leq(\cP')^*(X)=\cP^*(X)$. 
In particular,
$\mu(\{a,b\}),\mu(\{b,c\}),\mu(\{a,c\})\leq\frac{1}{2}$. Therefore,
\begin{equation}
\label{e:contr}
\mu(\{a,b\})+\mu(\{b,c\})+\mu(\{a,c\})\leq\frac{3}{2}.
\end{equation}
However, from standard properties of probability, it follows that
$$\mu(\{a,b\})+\mu(\{b,c\})+\mu(\{a,c\}) 
= 2\mu(\{a,b,c\})\\
> 2\times \frac{3}{4} = \frac{3}{2},$$
which contradicts (\ref{e:contr}). Therefore, $\mu$, and therefore
$\cP'$ cannot exist, and $\upm_\epsilon$ is not an upper probability
measure.
\eprf

\othm{t:fupm}
There exists constants $B_0,B_1,\ldots$ such that 
if $\alg$ is an algebra of subsets of
$\world$  and $\upm$ is a function
$\upm:\alg\rightarrow\cR$, then there exists a set $\cP$ of
probability measures such that $\upm=\cP^*$ if and only if $\upm$
satisfies the following properties:
\begin{quote}
\begin{itemize}
\item[\textbf{{\rm UPF1}}.] $\upm (\emptyset) = 0$,
\item[\textbf{{\rm UPF2}}.] $\upm (\world) = 1$,
\item[\textbf{{\rm UPF3}}.] 
for all integers $m, n, k \le B_{|\world|}$ and all sets $A_1, \ldots,
A_m$, if
$\om A_1, \ldots, A_m \cm$ is an $(n,k)$-cover of $(A,\world)$,
then $k+n\upm(A)\leq\sum_{i=1}^{m}\upm(A_i)$.
\end{itemize}
\end{quote}
\eothm
\prf
In view of Theorem \ref{t:upm}, we need only show that there exist
constant $B_0,B_1,\ldots$ such that a function $\upm$ satisfies
\textbf{UP3} iff it satisfies \textbf{UPF3}. Clearly, \textbf{UP3} always implies
\textbf{UPF3}, so it is sufficient to show that there exists $B_0,B_1,\ldots$
such that \textbf{UPF3} implies \textbf{UP3}. 

We need some terminology before
proceeding.  
An {\em exact $(n,k)$-cover\/} of $(A,\world)$ is a cover $C$ of $A$
with the property  
that every element of $A$ appears in exactly $n+k$ sets in $C$, and
every element of $\world-A$ appears in exactly $k$ sets in $C$. Thus,
while
an $(n,k)$-cover of  $(A,\world)$ can have many extra sets, as long as the 
sets cover $A$ at least $n+k$ times and $\world$ $k$ times, 
an exact cover has only the necessary sets, with the right total number
of elements. 
An exact
$(n,k)$-cover $C$ of $(A,\world)$ is \emph{decomposable} if there exists an
exact $(n_1,k_1)$-cover $C_1$ and an exact $(n_2,k_2)$-cover $C_2$ of
$(A,\world)$ such that $C_1$ and $C_2$ form a nontrivial partition of $C$,
with $n=n_1+n_2$ and $k=k_1+k_2$. Intuitively, an exact cover $C$ is
decomposable if it can be split into two exact covers. It follows
easily by induction that  for any exact $(n,k)$-cover, there exists a (not
necessarily unique) finite set of nondecomposable exact covers
$C_1,\ldots,C_m$, with $C_i$ an exact $(n_i,k_i)$-cover, such that 
the $C_i$'s a nontrivial partition of $C$ with $n=\sum_{i=1}^{m}n_i$ and
$k=\sum_{i=1}^{m}$. (If $C$ is itself nondecomposable, we can take
$m=1$ and $C_1=C$.) One can easily verify that if $C$ is an exact
$(n,k)$-cover of $(A,\world)$ and $C'\subseteq C$ 
is an exact
$(n',k')$-cover of $(A,\world)$
with $n' + k' < n+k$, then $C$ is decomposable.

The following lemma highlights the most important property of exact
covers from our perspective. It says that for any set $A\in\alg$, there
cannot be a ``large'' nondecomposable exact cover of  $(A,\world)$.

\lem
\label{l:covers}
There exists a sequence $B_1', B_2', B_3', \ldots$ such that for all $A
\subseteq \world$, every exact $(n,k)$-cover of $(A,\world)$ with
$n>B_{|\world|}'$ or $k>B_{|\world|}'$ is decomposable.  
\elem
\prf
It is clearly sufficient to show that for any finite $\world$
we can find a $B_{|\world|}$ with the required
properties. Fix a $\world$.  
Given $A \subseteq \world$, we first show that there exists $N_A$ such
that if $n > N_A$ or $k > N_A$, every exact $(n,k)$-cover of
$(A,\world)$ is decomposable.  Suppose for the sake of contradiction
that this is not the case.
This means that we can find an infinite
sequence $C_1, C_2,\ldots$ such that $C_i$ is a  nondecomposable exact
$(n_i,k_i)$-cover of $(A,\world)$, with either $n_1 < n_2 < \ldots$ or $k_1 <
k_2 < \ldots$.

To derive a contradiction, we use the following lemma,
known as Dickson's Lemma \cite{Dickson13}.
\begin{quote}
\lem 
\label{l:2}
   Every infinite sequence of $d$-dimensional vectors over the
   natural numbers contains a monotonically 
nondecreasing
subsequence in the
   pointwise ordering (where $x \leq y$ in the pointwise ordering iff
$x_i \leq y_i$ for all $i$). 
\elem
\prf 
It is straightforward to prove by induction on $k$ that if $k \le d$,
then every infinite sequence of vectors $x^1, x^2, \ldots$ contains a
subsequence $x^{i_1}, x^{i_2}, \ldots$ such that $x^{i_1}_j, x^{i_2}_j,
\ldots$ is a nondecreasing sequence of natural numbers for all $j \le
k$.  The base case is immediate from the observation that
every infinite sequence of natural numbers contains a 
nondecreasing subsequence.  For the inductive step, observe that if 
$x^{i_1}, x^{i_2}, \ldots$ is a subsequence such that $x^{i_1}_j,
x^{i_2}_j, \ldots$ is a nondecreasing sequence of natural numbers for
all $j \le k$, then the sequence $x^{i_1}_{k+1}, x^{i_2}_{k+1}, \ldots$
of natural numbers must have a nondecreasing subsequence.  This
determines a subsequence of the original sequence with the appropriate
property for all $j \le k+1$. \eprf
\end{quote}

Let $S_1,\ldots,S_{2^{|\world|}}$ be an arbitrary ordering of the
$2^{|\world|}$ 
subsets 
of $\world$. We can associate with any cover $C$ a $2^{|\world|}$-dimensional
vector $x^C =(x^C_1,\ldots,x^C_{2^{|\world|}})$, where $x^C_i$ is the
number of times 
the subset $S_i$ of $\world$ appears in the multiset
$C$. The key property of this association is that if $C'$ and $C$ are
multisets, then 
$C'\subseteq C$ iff $x^{C'}\leq x^C$ in the pointwise ordering.

Consider the sequence of vectors $x^{C_1},x^{C_2},\ldots$ associated
with the sequence $C_1,  C_2, \ldots$ of nondecomposable exact covers of $(A,
\world)$. By Lemma \ref{l:2}, there is a 
nondecreasing subsequence of vectors, $x^{C_{i_1}}\leq x^{C_{i_2}} \leq
\cdots$. But this 
means that $C_{i_1}\subseteq 
C_{i_2}\subseteq \cdots$. Since $n_1 < n_2 < \ldots$ or $k_1 < k_2 <
\ldots$, every cover in the chain must be distinct. But any pair of
exact covers in the chain is such that $C_i \subseteq C_{i+1}$,
meaning $C_{i+1}$ is decomposable, contradicting our
assumption. Therefore, there must exist an 
$N_A$ such that any exact $(n,k)$-cover of $A$ with $n> N_A$ or $k>N_A$ is
decomposable. 

Now define $B_{|\world|}' = \max\{N_A\sep A \subseteq \{1, \ldots,
|\world|\}\}$. 
It is easy to see that this choice works.
\eprf

To get the constants $B_1,B_2,\ldots$, let $B_N = 2NB_N'$, for
$N=1,2,\ldots$, where $B_N'$ is as in Lemma~\ref{l:covers}.
We now show that \textbf{UPF3} implies \textbf{UP3} with this choice of
$B_1,B_2,\ldots$.  Assume that \textbf{UPF3} holds.  Fix $\world$.
Suppose that 
$C = \om A_1, \ldots, A_m \cm$ is an $(n,k)$-cover of $(A,\world)$ with
$|C| = m$.  We want to show that
$k+n\upm(A)\leq\sum_{i=1}^{m}\upm(A_i)$. We proceed as follows.  

The first step is to show that,
without loss of generality, $C$ is an exact $(n,k)$-cover of
$(A,\world)$.   
Let $B_i$ consist of those states $s \in A_i$
such that either $s \in A$ and $s$ appears in more than $n+k$ sets in
$A_1, \ldots, A_{i-1}$ or $s \in \world - A$ and s appears in more than
$k$ sets in $A_1, \ldots, A_{i-1}$.  Let $A_i' = A_i - B_i$.  
Let $C' = \om A_1', \ldots, A_m' \cm$.  It is easy to check that $C'$ is
an exact $(n,k)$-cover of $(A,\world)$.  For if $s \in A$, then $s$
appears in exactly $n+k$ sets in $C'$ (it appears in $A'_j$ iff $A_j$ is
among the first $n+k$ sets in $C$ in which $s$ appeared) and, similarly,
if $s \in \world -A$, then $s$ appears in exactly $k$ sets in $C'$.
Clearly if \textbf{UP3} holds for $C'$, then it holds for $C$, since
$\upm(A_i') \le \upm(A_i)$ for $i = 1, \ldots, m$.  Thus, we can assume
without loss of generality that $C$ is an exact $(n,k)$-cover of $A$.

We can also assume without loss of generality that no set in $C$ is
empty (otherwise, we can simply remove the empty sets in $C$; the
resulting set is still an $(n,k)$-cover of $(A,\world)$).  There are now
two cases to consider.  If $\max(m,n,k)
\le B_{|\world|}$, the desired result follows from \textbf{UPF3}.  If
not, consider a decomposition of $C$ into multisets
$C_1, \ldots, C_p$, where $C_h$ is an exact $(n_h,k_h)$-cover of
$(A,\world)$ and is not further decomposable.  We claim that
$\max(|C_h|,n_h,k_h) \le B_{|\world|}$ for $h=1, \ldots, p$.  
If $n_h > B_{|\world|}$
or $k_h > B_{|\world|}$, then it is immediate from Lemma~\ref{l:covers}
that $C_h$ can be further 
decomposed, contradicting the fact that $C_h$ is not decomposable.  And if
$|C_h| >  B_{|\world|}$, then observe that 
$\sum_{X \in C_h} |X| \ge |C_h|$.  Since $|C_h| > B_{|\world|}
= 2|\world|B_{|\world|}'$, there must be some $s \in \world$ which
appears in at least $2B_{|\world|}'$ sets in $C_h$.  Since $C_h$ is an exact
$(n_h,k_h)$-cover, it follows that 
either $n_h > B_{|\world|}'$ or $k_h > B_{|\world|}'$.  But then, by
Lemma~\ref{l:covers}, $C_h$ is decomposable, again a contradiction. 

Now we can apply \textbf{UPF3} to each of $C_1, \ldots, C_k$ to get
\[\sum_{X\in C_h}\upm(X)-n_h\upm(A)\geq k_h.\] 
Since the $C_h$'s form a decomposition of $C$, we have 
\begin{eqnarray*}
 & & \sum_{h=1}^{p}\left(\sum_{X\in C_h}\upm(X)-n_h\upm(X)\right)\geq
\sum_{h=1}^{p}k_h \\
 & \Rightarrow & \sum_{h=1}^{p}\left(\sum_{X\in
C_h}\upm(X)\right)-\sum_{h=1}^{p}n_h\upm(A) \geq \sum_{h=1}^{p}k_h \\
 & \Rightarrow & \sum_{i=1}^{m}\upm(A_i)-(\sum_{h=1}^{p}n_h)\upm(A) \geq
\sum_{h=1}^{p}k_h
\end{eqnarray*}
By decomposition, $n=\sum_{h=1}^{p}n_h$ and $k=\sum_{h=1}^{p}k_h$, and 
therefore $\sum_{i=1}^{m}\upm(A_i)-n\upm(A)\geq k$, showing that
\textbf{UP3} holds, as desired.
\eprf

\opro{l:compl}
The formula $f$ is satisfiable in an upper probability structure iff the
inequality formula $\hat{f}$ has a solution.
Moreover, if $\hat{f}$ has a solution, then $f$ is satisfiable in an
upper probability structure with at most $2^{|f|}$ worlds.
\eopro
\prf
Assume first that $f$ is satisfiable. Thus there is some upper
probability structure $M=(\world,\alg,\cP,\pi)$ such that $M\models
f$. 
As 
in Section~\ref{s:compl}, let $p_1,\ldots,p_N$ be the
primitive propositions that appear in $f$, and let
$\rho_1,\ldots,\rho_{2^{2^N}}$ be some canonical listing of the
inequivalent formulas over $p_1, \ldots, p_N$. Without loss of 
generality, we assume that $\rho_1$ is equivalent to $\truep$, and
$\rho_{2^{2^N}}$ is equivalent to $\falsep$. 
Define the vector $x^*$ by letting $x^*_i=\cP^*(\eventM{\rho_i})$,
for $1\leq i\leq 2^{2^N}$. 
Since $M \models f$, it is immediate that $x^*$ is a solution to the
inequality formula $\overline{f}$.
Moreover, since $\rho_1 = \falsep$ and $\rho^{2^{2^N}} =
\truep$, it follows that 
$x^*_1=0$ (since $\cP^*(\eventM{\falsep})=\cP^*(\emptyset)=0$) and
$x^*_{2^{2^N}}=1$ (since $\cP^*(\eventM{\truep})=\cP^*(\world)=1$).  
Finally, consider a conjunct of $\hat{f}$ corresponding to an instance of
\axiom{L4}; suppose it has the form 
$x_{i_1} + \cdots x_{i_m} - n x_{i_{m+1}} \ge k$.
Since this conjunct appears in $\hat{f}$, it must be the case that
$(\rho_{i_{m+1}} \rimp\bigvee_{J\subseteq\{1,\ldots,m\},\,
|J|=k+n}\bigwedge_{J\subseteq\{1,\ldots,m\},\, j\in J}\rho_{i_j}) \land
(\bigvee_{|J|=k}\bigwedge_{j\in J}\rho_{i_j})$ is a propositional
tautology.  Thus, it follows that $\eventM{\rho_{i_1}}, \ldots,
\eventM{\rho_{i_m}}$ is an $(n,k)$-cover for 
$(\eventM{\rho_{i_{m+1}}},\eventM{\truep})$. 
It follows from UP3 that 
\[\cP^*(\eventM{\rho_{i_1}})+\cdots+\cP^*(\eventM{\rho_{i_m}})-n\cP^*(\eventM{\rho})\geq
k.\]    Thus, $x^*$ is a solution to the inequality formulas
corresponding to \axiom{L4}.  Hence, $x^*$ is a solution to $\hat{f}$.

For the converse, assume that $x^*$ is a solution to $\hat{f}$.
We construct an upper probability structure
$M=(S,E,\cP,\pi)$ such that $M\models f$ as follows. Let
$p_1,\ldots,p_N$ be the primitive propositions appearing in $f$. Let
$S=\{\delta_1,\ldots,\delta_{2^N}\}$ be the 
atoms over $p_1, \ldots, p_N$.
Let $E$ be the set of all subsets of $S$. 
As observed earlier, 
every propositional formula over $p_1, \ldots, p_n$ is equivalent to
a unique disjunction of atoms.  Thus, we can get a canonical collection
$\rho_1, \ldots, \rho_{2^{2^N}}$ of inequivalent formulas over $p_1,
\ldots, p_n$ by identifying each formula $\rho_i$ with a different
element of $E$,
where $\rho_1$ corresponds to the empty set and $\rho_{2^{2^N}}$
corresponds to all of $S$.
Define a set function $\upm$ by taking
$\upm(\{\delta_{i_1},\ldots,\delta_{i_j}\}) = x_i^*$ if $\rho_i$ is the
disjunction of the atoms  $\delta_{i_1},\ldots,\delta_{i_j}$.
Let $\pi(\delta)(\rho)=\true$ iff $\delta\rimp\rho$. 

It is now sufficient to show that $\upm$ is an upper probability (of a
set $\cP$ of probability measures), since then 
it is clear that $(S,E,\cP,\pi)\models f$ (since $x^*$ is a solution to $\hat{f}$, 
the system of inequalities derived from formula $f$). 
To do this, by Theorem~\ref{t:fupm}, it suffices to verify 
\textbf{UPF1}, \textbf{UPF2}, and \textbf{UPF3}, using $B_{2^N}$ in
\textbf{UPF3}, since $|S| = 2^N$. 
\begin{quote}
\begin{itemize}
\item[\textbf{UPF1}:] $\upm(\emptyset) = x^*_1 = 0$.
\item[\textbf{UPF2}:] $\upm(S) = x^*_{2^{2^N}} = 1$.
\item[\textbf{UPF3}:] Suppose that  $A$ and
$A_1,\ldots,A_m$ are in $E$ and satisfy
the premises of property \textbf{UPF3},
with $k, m, n \le B_{2^N}$.
Let $\rho_{i_1},\ldots,\rho_{i_m}, \rho_{i_{m+1}}$ 
be the canonical formulas corresponding to 
$A_1,\ldots,A_m,A$, respectively. Clearly,
$A\subseteq\bigcup_{J\subseteq\{1,\ldots,m\},\, |J|=k+n}\bigcap_{j\in  
J}A_{i_j}$ iff 
$\rho_{i_{m+1}}\rimp
\bigvee_{J\subseteq\{1,\ldots,m\},\,
|J|=k+n}\bigwedge_{j\in J}\rho_{i_j}$ 
is a propositional tautology
and 
similarly
$\world\subseteq\bigcup_{J\subseteq\{1,\ldots,m\},\,|J|=k}\bigcap_{j\in
J}A_{i_j}$ iff 
$\bigvee_{J\subseteq\{1,\ldots,m\},\,|J|=k}\bigwedge_{j\in J}\rho_{i_j}$
is a propositional tautology. 
Thus, $\sum_{j=1}^{m}x_{i_j} - x_{i_{m+1}}\geq k$ is one of the
inequality formulas in $\hat{f}$.  Thus, it follows that
$\sum_{j=1}^{m}x^*_{i_j} - x^*_{i_{m+1}}\geq k$, as desired.
By our definition of $\upm$, we therefore have
$k+n\upm(A)\leq\sum_{i=1}^{m}\upm(A_i)$, and so \textbf{UPF3} holds. 
~~\eprf
\end{itemize}
\end{quote}

\othm{t:smallmodel}
Suppose $f$ is a likelihood formula that is satisfied in some upper
probability structure. Then $f$ is satisfied in a structure
$(\world,\alg,\cP,\pi)$, where $|\world| \le |f|^2$, $\alg =
2^{\world}$ (every subset of $\world$ is measurable), $|\cP|\leq|f|$,
$\mu(w)$ is a rational
number such that $||\mu(w)||$ is
$O(|f|^2||f||+|f|^2\log(|f|))$ for every world $w \in \world$ and
$\mu \in \cP$,
and $\pi(w)(p) = \false$ for every world $w \in \world$ and every primitive
proposition $p$ not appearing in $f$. 
\eothm

\prf
The first step in the proof involves showing that
if $\cP$ is a set of probability measures defined on an algebra $\alg$
of a finite space $\world$, we can assume without loss of generality
that for each set $X \in \alg$, there is a probability measure $\mu_X
\in \cP$ such that $\mu_X(X) = \cP^*(X)$ (rather than $\cP^*(X)$ just
being the sup of $\mu(X)$ for $\mu \in \cP$).

\lem\label{achievemax}
Let $\cP$ be a set of probability measures defined on an algebra $\alg$
over a finite set $\world$.  Then there exists a set $\cP'$ of
probability measures such that, for each $X \in \alg$, $\cP^*(X) =
(\cP')^*(X)$; moreover, there is a probability measure $\mu_X \in \cP'$
such that $\mu_X(X) = \cP^*(X)$.  In addition, for any interpretation
$\pi$, if
$M = (\world, \alg, \cP,\pi)$ and $M = (\world,\alg,\cP',\pi)$, then for
all likelihood formulas $f$, $M \sat f$ iff $M' \sat f$.
\elem

\prf  Since $\alg$ is finite, to show that $\cP'$ exists,
it clearly suffices to show that, for each $X \in \alg$, there is
a probability measure $\mu_X$ such that $\mu_X(X) = \cP^*(X)$ and,
if $\cP' = \cP \union \{\mu_X\}$, then $\cP^*(Y) = (\cP')^*(Y)$ for all
$Y \in \alg$.  

Given $X$, if there exists $\mu \in \cP$ such that $\mu(X) = \cP^*(X)$,
then we are done.  Otherwise, we construct a sequence $\mu_1, \mu_2,
\ldots$ of probability measures in $\cP$ such that $\lim_i \mu_i(X) =
\cP^*(X)$ and, for all $Y \in \alg$, the sequence $\mu_i(Y)$ converges
to some limit.  
Let $X_1, \ldots, X_n$ be an enumeration of the sets in $\alg$, with
$X_1 = X$.  We inductively construct a sequence of measures
$\mu_{m1}, \mu_{m2}, \ldots$ in $\cP$ for $m \le n$ such that $\mu_{mi}(X_j)$
converges to a limit for $i \le k$ and $\lim_{i \rightarrow \infty}
\mu_{mi}(X) = \cP^*(X)$.  
For $m=1$, we know there must be a sequence $\mu_{11},
\mu_{12}, \ldots$ of measures in $\cP$ such that 
$\mu_{1i}(X)$ converges to $\cP^*(X)$.  For the inductive step, if $m <
n$, suppose we have constructed an appropriate sequence $\mu_{m1},
\mu_{m2}, \ldots$.  Consider 
the sequence of real numbers $\mu_{mi}(X_{m+1})$.
Using the Bolzano-Weierstrass theorem
\cite{Rudin76} (which says that every sequence of real numbers has a
convergent subsequence), this sequence  has a convergent subsequence.  Let
$\mu_{(m+1)1}, \mu_{(m+1)2}, \ldots$ be the subsequence of $\mu_{m1},
\mu_{m2}, \ldots$ which generates this convergent subsequence.
This sequence of probability measures clearly has all the required
properties.  This completes the inductive step.

Define $\mu_X(Y) = \lim_{i \rightarrow \infty}\mu_{ni}(Y)$.  It is easy
to check that that $\mu_X$ is indeed a probability measure, that $\mu_X(X) =
\cP^*(X)$, and if $\cP' = \cP \union \{\mu_X\}$, that $\cP^*(Y) =
(\cP')^*(Y)$ for all $Y \in \alg$.  This shows that an appropriate set
$\cP'$ exists.

Now, given $\pi$, let $M = (\world, \alg, \cP,\pi)$ and $M' =
(\world,\alg,\cP',\pi)$.  A straightforward induction on the structure
of $f$ shows that $M \sat f$ iff $M' \sat f$.  For the base case:
\begin{eqnarray*}
& & (\world,\alg,\cP,\pi)\models a_1l(\phi_1)+\cdots+a_n l(\phi_n)\geq
a \\ & \siff & a_1\cP^*(\eventM{\phi_1})+\cdots+a_n\cP^*(\eventM{\phi_n})\geq a \\
 & \siff &
a_1(\cP')^*(\eventMp{\phi_1})+\cdots+a_n(\cP')^*(\eventMp{\phi_n})\geq
a\\ 
 & \siff & (\world,\alg,\cP',\pi)\models a_1l(\phi_1)+\cdots+a_n
l(\phi_n)\geq a.
\end{eqnarray*}
The others cases are trivial. 
\eprf

\COMMENTOUT{
Our first step consists of showing that if $f$ is satisfied in some
upper probability structure with a finite set of worlds, it is satisfied in an upper probability 
structure $M=(\world,\alg,\cP,\pi)$ 
where $|\cP|$ is small.

\lem
\label{l:smallmodel}
Suppose $f$ is a likelihood formula that is satisfied in some upper
probability structure $M = (\world, \alg, cP,\pi)$ such that $\world$ is
finite.   
Then there is a set $\cP'$ such that $|\cP'| \le |f|$
and $f$ is satisfied in $(\world,\alg,\cP',\pi)$.
\elem

\prf 
Let $\cP''$ be the set of probability measures
(guaranteed to exist by Lemma \ref{achievemax}) such that $\cP^*(Y) =
(\cP'')^*(Y)$ for all $Y \in \alg$.  Let $\phi_1,\ldots,\phi_k$ be the
propositional formulas appearing in $f$. Clearly, $k\leq |f|$.  
Let $\mu_1, \ldots, \mu_k$ be probability measures in $\cP''$ such that 
$\mu_i(\eventM{\phi_i})=(\cP'')^*(\eventM{\phi_i})$, and let
$\cP'=\{\mu_i\}_{i\leq k}$.  Note that $(\cP')^*(\eventM{\phi_i}) =
\cP^*(\eventM{\phi_i})$ for $i = 1, \ldots, k$.

We claim $(\world,\alg,\cP',\pi)\models f$. 
We proceed by structural induction on $f$. For the base case:
\begin{eqnarray*}
& & (\world,\alg,\cP,\pi)\models a_1l(\phi_1)+\cdots+a_n l(\phi_n)\geq
a \\ & \siff & a_1\cP^*(\eventM{\phi_1})+\cdots+a_n\cP^*(\eventM{\phi_n})\geq a \\
 & \siff &
a_1(\cP')^*(\eventM{\phi_1})+\cdots+a_n(\cP')^*(\eventM{\phi_n})\geq
a\\
 & \siff & (\world,\alg,\cP',\pi)\models a_1l(\phi_1)+\cdots+a_n
l(\phi_n)\geq a.
\end{eqnarray*}
The others cases are trivial. 
\eprf}%

Just as in FHM, to prove Theorem~\ref{t:smallmodel}, we make use
of the following lemma which can be derived from Cramer's rule
\cite{Shores99} and simple estimates on the size of the determinant (see
also \cite{Chv} for a simpler variant): 
\lem  
\label{l:chv}
If a system of $r$ linear equalities and/or inequalities with integer
coefficients each of length at most $l$ has a nonnegative solution,
then it has a nonnegative solution with at most $r$ entries positive,
and where the size of each member of the solution is
$O(rl+r\log(r))$. 
\elem

Continuing with the proof of Theorem~\ref{t:smallmodel}, 
suppose that $f$ is satisfiable in an upper probability structure.  By
Proposition~\ref{l:compl}, the system $\hat{f}$ of equality formulas has
a solution, so 
$f$ is satisfied in
a upper probability structure with a finite state space.  Thus, by
Lemma~\ref{achievemax}, $f$ is satisfied in a structure
$M=(\world,\alg,\cP,\pi)$ such that for all $X \in \alg$, there exists
$\mu_X \in \cP$ such that $\mu_X(X) = \cP^*(X)$.

As in the completeness proof, we can write $f$ in disjunctive normal
form. Each disjunct $g$ is a conjunction of at most $|f|-1$ basic
likelihood formulas and their negations.  
Since $M \sat f$, there must be some disjunct $g$ such that $M \sat g$.
Suppose that $g$ is the  
conjunction of $r$ basic likelihood formulas and $s$ negations of
basic likelihood formulas. Let $p_1,\ldots,p_N$ be the primitive
formulas appearing in $f$. Let $\delta_1,\ldots,\delta_{2^N}$ be the
atoms over $p_1, \ldots, p_N$.
As in the proof of completeness, we
derive a system of equalities and inequalities from $g$. It is a 
slightly more complicated
system, however. 
Recall that each propositional formula over $p_1, \ldots, p_N$ is a
disjunction of atoms.  Let $\phi_1, \ldots, \phi_k$ be the propositional
formulas that appear in $g$.  Notice that $k < |f|$ (since there are
some symbols in $f$, such as the coefficients, that are not in the
propositional formulas).  The system of
equations and inequalities we 
construct involve variables $x_{ij}$, where $i = 1, \ldots, k$ and $j = 1,
\ldots, 2^N$. 
Intuitively, $x_{ij}$ represents
$\mu_{\eventM{\phi_i}}(\eventM{\delta^j})$, where 
$\mu_{\eventM{\phi_i}} \in \cP$ is such that 
$\mu_{\eventM{\phi_i}}(\eventM{\phi_i}) = \cP^*(\eventM{\phi_i})$.
Thus, the system includes 
$k<|f|$
equations of the following form,
\[x_{i1} + \cdots + x_{i2^N} = 1,\]
for $i = 1, \ldots, k$.
Since $\mu_{\eventM{\phi_i}}(\eventM{\phi_i}) \ge \mu(\eventM{\phi_i})$
for all $\mu \in \cP$, if $E_i$ is the subset of $\{1, \ldots, 2^N\}$
such that 
$\phi_i=\bigvee_{j\in E_i}\delta_j$, the system includes $k^2 - k$
inequalities of the form 
\[ \sum_{j\in E_i}x_{ij} \geq \sum_{j\in E_i}x_{i'j}, \]
for each pair $i$, $i'$ such that $i \ne i'$.
For each conjunct in $g$ of the form $\theta_1 l(\phi_1)+\cdots+\theta_n
l(\phi_k)\geq \alpha$, there is a corresponding inequality where,
roughly speaking, we replace $l(\phi_i)$ by
$\mu_{\eventM{\phi_i}}(\eventM{\phi})$.%
\footnote{For simplicity here, we are implicitly assuming that each of
the formulas $\phi_i$ appears in each conjunct of $g$.  This is without
loss of generality, since if $\phi_i$ does not appear, we can put it in,
taking $\theta_i = 0$.}
Since $\mu_{\eventM{\phi_i}}$ corresponds to $\sum_{j \in E_i} x_{ij}$, 
the appropriate inequality is 
\[\sum_{i=1}^k \theta_i\sum_{j\in E_i}x_{ij} \geq \alpha.\]
Negations of such formulas correspond to a negated inequality formula;
as before, this is equivalent to a formula of the form
\[-(\sum_{i=1}^k \theta_i\sum_{j\in E_i}x_{ij}) > -\alpha.\]
Notice that there are at most $|f|$ inequalities corresponding to the
conjuncts of $g$.  Thus, altogether, there are at most $k(k-1) + 2|f| < |f|^2$ 
equations and inequalities in the system (since $k < |f|$).
We know that the system has a nonnegative solution (taking $x_j^i$ to be
$\mu_{\eventM{\phi_i}}(\eventM{\delta^j})$).  
It follows from Lemma \ref{l:chv} that
the system has a solution
$x^*=(x^*_{11},\ldots,x^*_{12^N},\ldots,x^*_{k1},\ldots,x^*_{k2^N})$  
with $t\leq |f|^2$ entries positive, and with each entry of size $O(|f|^2||f||+|f|^2\log(|f|))$. 

We use this solution to 
construct a small structure satisfying 
the formula
$f$. Let 
$I=\{i \sep x^*_{ij}~\mbox{is positive, for some $j$}\}$; suppose that $I =
\{i_1, \ldots, i_{t'}\}$, for some $t' \le t$.
Let $M=(S,E,\cP,\pi)$ where $S$ has $t'$ states, say
$s_1,\ldots,s_{t'}$, and $E$ consists of all subsets of $S$. Let
$\pi(s_h)$ be the truth assignment corresponding to the formula
$\delta_{i_h}$, 
that is, $\pi(s_h)(p) = \true$ if and only if $\delta_{i_h}\rimp p$ 
(and where $\pi(s_h)(p)=\false$ if $p$ does not appear 
in $f$). Define $\cP=\{\mu_j\sep 1\leq i\leq k\}$, where $\mu_j(s_h) = x^*_{i_hj}$.
It is clear from the construction that $M\models f$. Since
$|\cP|=k < |f|$, $|S|=t'\leq t\leq |f|^2$ and $\mu_j(s_h) = x^*_{i_hj}$,
where, by construction, the size of $x^*_{i_hj}$ is
$O(|f|^2||f||+|f|^2\log(|f|))$,  the theorem follows. 
\eprf

\section{Proof of the Characterization of Upper Probabilities}
\label{a:upm}

To make this paper self-contained, in this appendix we  give a proof
of Theorem
\ref{t:upm}. The proof we give is essentially that of Anger and
Lembcke \citeyear{Anger85}. 
Walley~\citeyear{Walley91} gives an alternate proof along somewhat
similar lines.
Note that the functional
$\tg$ we define in our proof corresponds to the construction in
Walley's Natural Extension Theorem, which is needed in his version of
this result. 

\othm{t:upm}
Suppose that $\world$ is a set, $\alg$ is an algebra of subsets of
$\world$, and $\upm:\alg\rightarrow\cR$. Then there
exists a set $\cP$ of probability measures with $\upm=\cP^*$ if and
only if $\upm$ satisfies the following three properties: 
\begin{quote}
\begin{itemize}
\item[\textbf{{\rm UP1}}.] $\upm (\emptyset) = 0$,
\item[\textbf{{\rm UP2}}.] $\upm (\world) = 1$,
\item[\textbf{{\rm UP3}}.] for all integers $m,n,k$ and all subsets
$A_1, \ldots, A_m$ in $\alg$, if $\om A_1, \ldots, A_m \cm$ is an
$(n,k)$-cover of $(A,\world)$, then
$k+n\upm(A)\leq\sum_{i=1}^{m}\upm(A_i)$.
\end{itemize}
\end{quote}
\eothm
\prf
The ``if'' direction of the characterization is
straightforward. Given $\cP=\{\mu_i\}_{i\in I}$ a set of probability
measures, we show $\cP^*$ satisfies \textbf{UP1}-\textbf{UP3}. 
\begin{quote}
\begin{itemize}
\item[\textbf{UP1}:] $\cP^*(\emptyset) = \sup\{\mu_i(\emptyset)\} = \sup \{0\} = 0$
\item[\textbf{UP2}:] $\cP^*(\world) = \sup\{\mu_i(\world)\} = \sup\{1\} = 1$
\item[\textbf{UP3}:] Given $A_1,\ldots,A_m$ and $A$ such that
$A\subseteq\bigcup_{J\subseteq\{1,\ldots,m\},|J|=k+n}\bigcap_{j\in J}A_{i_j}$ 
and
$\world\subseteq\bigcup_{J\subseteq\{1,\ldots,m\},|J|=k}\bigcap_{j\in J}A_{i_j}$, 
then for any $i$ we have $k\mu_i(\world) + n \mu_i(A) \leq
\sum_{j=1}^{m}\mu_i(A_j)$, that is
$k+n\mu_i(A)\leq\sum_{j=1}^{m}\mu_i(A_j)\leq\sup_i\{\sum_{j=1}^{m}\mu_i(A_j)\}\leq\sum_{j=1}^{m}\sup_i\{\mu_i(A_j)\}=\sum_{j=1}^{m}\cP^*(A_j)$.
But $\sup_i\{k+n\mu_i(A)\} = k+n\sup_i\{\mu_i(A)\} = k+n\cP^*(A)$, so $k+n\cP^*(A) \leq
\sum_{j=1}^{m}\cP^*(A_j)$, as required.
\end{itemize}
\end{quote}

As for the ``only if'' direction, we first prove a general lemma relating the
problem to the Hahn-Banach Theorem. 
Some general definitions are needed.
Suppose that we are given a space $W$ and an algebra $\cF$ of subsets
of $W$. Let $\cK$ be the vector space generated by the indicator
functions $1_X$ defined by 
\[1_X(x) = \left\{\begin{array}{ll}0 & \mbox{if $x\not\in X$} \\ 1 &
\mbox{if $x\in X$,}\end{array}\right.\]
for $X\in\cF$. A \emph{sublinear functional on \cK} is a mapping
$c:\cK\rightarrow\cR$ such that $c(\alpha h)=\alpha c(h)$ for
$\alpha\geq 0$ and $c(h_1+h_2)\leq c(h_1)+c(h_2)$ for all $h_1,h_2$. A 
sublinear functional is \emph{increasing} if $h\geq 0$ implies
$c(h+h')\geq c(h')$ for all $h'\in\cK$. 
The following result is a formulation of the well-known Hahn-Banach
Theorem (see, for example, \cite{Conway90a}).

\nthm{(Hahn-Banach)}
Let $\cK$ be a vector space over $\cR$, and let $g$ be a
sublinear functional on $\cK$. If $\mathcal{M}$ is a linear
subspace in $\cK$ and $\lambda:\mathcal{M}\rightarrow\cR$ is a
linear functional such that $\lambda(x)\leq g(x)$ for all $x$ in
$\mathcal{M}$, then there is a linear functional
$\lambda':\cK\rightarrow\cR$ such that $\lambda'|_{\mathcal{M}}=\lambda$ and $\lambda'(x)\leq g(x)$ 
for all $x$ in $\cK$. 
\enthm

\lem
\label{l:step1}
Let $g:\cF\rightarrow[0,1]$ be such that $g(W)=1$ and suppose that there is an
increasing sublinear functional $\tg$ on $\cK$ such that
\begin{enumerate}
\item $\tg(1_K)=g(K)$ for $K\in\cF$;
\item $\tg(h)\leq 0$ if $h\leq 0$;
\item $\tg(-1)\leq-1$ (where $\tg(\alpha)$ is identified with $\tg(\alpha
1_W)$).
\end{enumerate}
Then $g$ is an upper probability measure.
\elem

\prf
We show that $g$ is an upper probability by exhibiting a set
$\{\mu_X \sep X\in\alg\}$ of probability measures, with the property that 
$\mu_X(X)=g(X)$ and $\mu_X(Y)\leq g(X)$ for $Y\ne X$.   Each
probability measure $\mu_X$ is constructed through an application of
the Hahn-Banach Theorem.

Given $X\in\cF$, define the linear functional $\lambda$ on the
subspace generated by $1_X$ by $\lambda(\alpha 1_X)=\alpha \tg(1_X)$. We 
claim that $\lambda(h)\leq \tg(h)$ for all $h$ in the subspace. Since
the elements of the subspace have the form $\alpha 1_X$, there are two 
cases to consider: $\alpha\geq 0$ and $\alpha<0$. If $\alpha\geq 0$,
then $\lambda(\alpha 1_X)=\alpha \tg(1_X)=\tg(\alpha
1_X)$, since $\tg$ is sublinear. Moreover, $0=\tg(0)=\tg(-1_X+1_X)\leq
\tg(-1_X)+\tg(1_X)$, so $\tg(-1_X)\geq-\tg(1_X)$. Thus, if $\alpha>0$, then
\[\lambda(-\alpha 1_X)=-\alpha \tg(1_X)\leq\alpha \tg(-1_X)=\tg(-\alpha
1_X).\] 

Now, by the Hahn-Banach Theorem, we can extend $\lambda$ to a linear
functional $\lambda'$ on all of $\cK$ such that $\lambda'(h)\leq \tg(h)$
for all $h$. We claim that (a) $\lambda'(1_Y)\geq 0$ for all $Y\in\cK$ 
and (b) $\lambda'(1)=1$. For (a), note that $\lambda'(-1_Y)\leq
\tg(-1_Y)\leq 0$ by assumption, so 
$\lambda'(1_Y)\geq 0$. For (b),
note that $\lambda'(1)\leq \tg(1)=g(W)=1$ and that
$\lambda'(1)=-\lambda'(-1)\geq-\tg(-1)\geq 1$ (since $\tg(-1)\leq -1$, by
assumption). 

Define $\mu_X(Y)=\lambda'(1_Y)$. Since $\lambda'(1_W)=1$,
$\mu_X(W)=1$. If $Y$ and $Y'$ are disjoint, it is immediate
from the linearity of $\lambda$ that $\mu_X(Y\cup
Y')=\mu_X(Y)+\mu_X(Y')$. By construction, $\mu_X(Y)\leq
\tg(1_Y)=g(Y)$ for any $Y\not= X$, and $\mu_X(X)=\lambda(1_X)=\tg(1_X) 
= g(X)$. Bottom line: there is a probability
measure $\mu_X$ dominated by $g$ such that $\mu_X(X)=g(X)$. 

Take $\cP=\{\mu_X \sep X\in\alg\}$. Since for any $X$ we have that
$\mu_X(X)=g(X)$ and $\mu_X(Y)\leq g(X)$ (if $Y\not= X$), we have
$\cP^*(X) = \mu_X(X) = g(X)$. Therefore, 
$g = \cP^*$.
\eprf

The main result follows by showing how to construct,  from a function
$\upm$ satisfying the properties of Theorem \ref{t:upm}, a
sublinear functional $c$ on $\cK$ with the required properties. 

Suppose that $g:\alg\rightarrow\cR$ is a function satisfying
\textbf{UP1}-\textbf{UP3}.  
As we show in the discussion after Theorem~\ref{t:upm} in the text,
\textbf{UP1}-\textbf{UP3} show that the range of $g$ is in fact
$[0,1]$. 
Since $g$ satisfies \textbf{UP3}, if 
$\om K_1,\ldots,K_m \cm$ is an $(n,k)$-cover of $(K,\world)$,
we have $k+n g(K)\leq\sum_{i=1}^{m}K_i$. This is equivalent to 
saying that $k+n 1_K  \leq
\sum_{i=1}^{m}1_{K_i}$.  Hence, for all $K_1,\ldots,K_m$ such that
$k+n 1_K\leq\sum_{i=1}^m1_{K_i}$, we have $k+n g(K)\leq\sum_{i=1}^{m}
g(K_i)$, or equivalently 
\begin{equation}\label{eq.UP3}
-\frac{k}{n}
+\frac{1}{n}\sum_{i=1}^{m} g(K_i)\geq g(K). 
\end{equation}

This observation motivates the following definition of the functional
$\tg:\cK\rightarrow
\cR \union \{-\infty,\infty\}$:
\[\tg(h) =
\inf\left\{-\frac{k}{n}+\frac{1}{n}\sum_{i=1}^{m}g(K_i) : m,n,k\in\cN, 
\, m, n > 0, \, 
K_1,\ldots,K_m\in\cF,
\,-\frac{k}{n}+\frac{1}{n}\sum_{i=1}^{m}1_{K_i}\geq 
h\right\}.\]
Our goal now is to show that $\tg$ satisfies the conditions of
Lemma~\ref{l:step1}.  
\begin{itemize}
\item It is almost immediate from the definitions that $\tg$ is
increasing: if $h \ge 0$ and
$-\frac{k}{n}+\frac{1}{n}\sum_{i=1}^{m}1_{K_i}\geq  
h + h'$, then $-\frac{k}{n}+\frac{1}{n}\sum_{i=1}^{m}1_{K_i}\geq 
h'$.  
\item To see that $\tg$ is sublinear, note that it is easy to see using
the properties of inf that 
$\tg(h_1 + h_2) \le \tg(h_1) + \tg(h_2)$.  
To show that 
$\tg(\alpha h) = \alpha \tg (h)$
for $\alpha \ge 0$,
first observe that the definition of $\tg$ is equivalent to 
\[\inf\left\{-\beta+\sum_{i=1}^{m}\beta_i g(K_i) : 
m\in\cN, \beta, \beta_i\in\cR_{+}, 
K_1,\ldots,K_m\in\cF\, -\beta+\sum_{i=1}^{m}\beta_i
1_{K_i}\geq h\right\}.\]
Consider first the case $\alpha>0$. Then 
\begin{eqnarray*}
\tg(\alpha h) & = & \inf\left\{-\beta+\sum_{i=1}^{m}\beta_i
g(K_i) \sep -\beta+\sum_{i=1}^{m}\beta_i 
1_{K_i}\geq \alpha h\right\}\\
& = & \inf\left\{-\beta+\sum_{i=1}^{m}\beta_i 
g(K_i) \sep -\frac{\beta}{\alpha}+\frac{1}{\alpha}\sum_{i=1}^{m}\beta_i
1_{K_i}\geq h\right\} \\
& = &
\alpha\inf\left\{-\frac{\beta}{\alpha}+\frac{1}{\alpha}\sum_{i=1}^{m}\beta_i 
g(K_i) \sep -\frac{\beta}{\alpha}+\frac{1}{\alpha}\sum_{i=1}^{m}\beta_i
1_{K_i}\geq h\right\} \\
& = & \alpha\tg(h).
\end{eqnarray*}
For $\alpha=0$,   it is clear from the definition of $\tg$ that
$\tg(1_{\emptyset})\leq g(\emptyset)$. From (\ref{eq.UP3}) it follows
that $\tg(1_{\emptyset})\geq g(\emptyset)$, and hence
$\tg(0)=\tg(1_{\emptyset})=g(\emptyset)=0$. 
\item It is immediate from 
the definition of $\tg$ that $\tg(1_{K})\leq g(K)$ for
$K\in\cF$; the fact that $\tg(1_{K}) = g(K)$  now follows from
(\ref{eq.UP3}).   
\item It is 
immediate from the definition that $\tg(-1)\leq -1$.  
\item If $h\leq 0$, then $-h\geq 0$; since $\tg$ is  
increasing, $\tg(h)\leq\tg(-h + h)=\tg(0)$, and since $\tg$ is
sublinear, $\tg(0)=0$.
\end{itemize}
Since the conditions of
Lemma \ref{l:step1} are satisfied, $g$ is an upper probability measure.
\eprf

\bibliographystyle{theapa}
\bibliography{riccardo,z,joe}

\end{document}